\documentclass[runningheads]{llncs}
\usepackage{graphicx}
\usepackage{amsmath}
\usepackage{amsfonts}
\usepackage[table,xcdraw]{xcolor}
\usepackage{booktabs}
\usepackage{float}
\usepackage{subcaption}
\usepackage{bbm}
\usepackage{url}
\usepackage{hyperref}
\usepackage{wrapfig}
\usepackage{amssymb}
\usepackage{amsmath}
\usepackage{pdflscape}
\usepackage{booktabs}
\usepackage{url}
\usepackage{bbm}
\usepackage{xcolor}
\usepackage{multirow}
\usepackage{float}
\usepackage{hyperref}

\begin{document}
\setlength{\abovedisplayskip}{10pt}
\setlength{\belowdisplayskip}{10pt}
\title{Investigating the Impact of Balancing, Filtering, and Complexity on Predictive Multiplicity: \\A Data-Centric Perspective}
\titlerunning{Investigating the data preprocessing on predictive multiplicity}
%
\author{Mustafa Cavus\inst{1}\orcidID{0000-0002-6172-5449} \and
Przemysław~Biecek\inst{2,3}\orcidID{0000-0001-8423-1823}}
\authorrunning{Cavus and Biecek}
%
\institute{Eskisehir Technical University, Department of Statistics, Turkiye\\
\email{mustafacavus@eskisehir.edu.tr} 
\and Warsaw University of Technology, Faculty of Mathematics and Information Science, Poland \and University of Warsaw, Faculty of Mathematics, Informatics and
Mechanics, Poland}
\maketitle              

\begin{abstract}
The Rashomon effect presents a significant challenge in model selection. It occurs when multiple models achieve similar performance on a dataset but produce different predictions, resulting in predictive multiplicity. This is especially problematic in high-stakes environments, where arbitrary model outcomes can have serious consequences. Traditional model selection methods prioritize accuracy and fail to address this issue. Factors such as class imbalance and irrelevant variables further complicate the situation, making it harder for models to provide trustworthy predictions. Data-centric AI approaches can mitigate these problems by prioritizing data optimization, particularly through preprocessing techniques. However, recent studies suggest preprocessing methods may inadvertently inflate predictive multiplicity. This paper investigates how data preprocessing techniques like balancing and filtering methods impact predictive multiplicity and model stability, considering the complexity of the data. We conduct the experiments on 21 real-world datasets, applying various balancing and filtering techniques, and assess the level of predictive multiplicity introduced by these methods by leveraging the Rashomon effect. Additionally, we examine how filtering techniques reduce redundancy and enhance model generalization. The findings provide insights into the relationship between balancing methods, data complexity, and predictive multiplicity, demonstrating how data-centric AI strategies can improve model performance.

\end{abstract}
\newpage
\section{Introduction}

The traditional model selection methods typically involve training multiple models in various machine learning (ML) pipelines and choosing the one with the highest accuracy. However, the presence of the Rashomon effect makes this strategy ineffective. It highlights situations where multiple models perform similarly well on a dataset \cite{Breiman_2001}, yet may yield divergent predictions for the same observations leading to predictive multiplicity \cite{Marx_et_al_2020}. The variation of predictions can disparately impact individual samples. The traditional way of model selection may target and cause systemic harm to specific individuals by excluding them from favorable outcomes in high-stakes domains. This is observed because the datasets are often underspecified and can be modeled by multiple models that each describe the data equally well; however, the data offers no further evidence to prefer one over another \cite{Lee_et_al_2023}. Even all of this happens silently because high accuracy hides it \cite{Long_et_al_2024, Renard_et_al_2024}. Because researchers commonly select the most accurate model from the set of models they have trained and deploy it, but do not focus on predictive multiplicity.

The ML pipeline is organized according to the factors that complicate the decision-making processes of the models. For example, one of the most persistent factors is class imbalance in classification tasks, where models exhibit a bias toward the majority class, resulting in poor predictive performance for the minority class \cite{Luengo_et_al_2011}. The other one is the existence of irrelevant variables increases the dimension of data, which can raise the model's learning difficulty and lead to lower prediction accuracy \cite{Hou_et_al_2023}. Model performance can additionally suffer due to factors beyond class imbalance, including data complexity and irregularity, emphasizing the challenge \cite{Santos_et_al_2023}. Even with severe class imbalance, standard classifiers can achieve good results when the data is not complex \cite{Fernandez_et_al_2018}. While model-centric approaches can address this issue, data-centric strategies offer more practical solutions. 

In model-centric AI, researchers focus on developing more efficient models to improve AI performance, often neglecting the importance of data. This leads to the oversight of data-related issues, such as class imbalance, missing observations, incorrect labels, and outliers, which can adversely affect model performance \cite{Zha_et_al_2023}. The data-centric perspective prioritizes data optimization methods over ML models and hyperparameter optimization to enhance model performance. In this context, data-centric AI aims to improve data quality continuously. Data preprocessing efforts toward this goal constitute a crucial part of data-centric AI \cite{Singh_2023}.

It is quite easy to apply data preprocessing methods in the ML pipeline, also known as data-centric AI solutions, to solve these problems in classification problems. Balancing methods are used to eliminate class imbalance and have been preferred in several domains. However, in recent years, many side effects of these methods have been revealed \cite{Stando_et_al_2024,Patil_et_al_2020,Alarab_and_Prakoonwit_2022,Goorbergh_et_al_2022,Simson_et_al_2023}, and the emergence of predictive multiplicity due to the inflation in the Rashomon effect \cite{Cavus_and_Biecek_2024}. This underscores the need for a deeper understanding of model behavior during the model selection process, as performance alone is insufficient for trustworthy ML. On the other hand, it provides an opportunity to enhance the model by emphasizing different variables while relying solely on accuracy for model choice, which may appear flawed under the Rashomon effect. It thus facilitates a major transformation in how models are assessed, moving beyond accuracy. Nevertheless, while there has been a growing research interest, it remains underexplored relative to its worth \cite{Biecek_and_Samek_2024}.

In this paper, we focused on the effect of balancing methods on predictive multiplicity concerning the complexity of data. Because the correlation between the imbalance ratio and model performance is low, it is an unreliable indicator of problem complexity \cite{Gottcke_et_al_2024}. Thus, we show the importance of selecting a strategy that is concerned with the complexity of data. Regarding this aim, we are first considering ten commonly used balancing methods to evaluate their effects on predictive multiplicity. Then, we propose elimination strategies against the predictive multiplicity using filtering methods based on correlation and significance tests for variable selection. It aims to improve the model's predictive performance by removing irrelevant variables. They also seek to simplify the model, reduce training time, enhance generalization by mitigating overfitting, and address the problem of high dimensionality \cite{Chen_et_al_2020}. Filtering methods may be vital in lowering predictive multiplicity by selecting statistically significant variables that minimize redundancy. While the correlation test aims to remove highly correlated variables to enhance stability, the significance test concentrates on selecting variables with distinct class distributions. These approaches complement balancing methods by ensuring that the input data fed to the models is balanced and devoid of confounding influences. The interplay between balancing and filtering methods remains underexplored, particularly their joint impact on predictive multiplicity across different data complexities. 

This paper fills this gap by investigating the Rashomon effect of balancing methods on models regarding predictive multiplicity. We conduct experiments on 21 real-world imbalanced datasets using balancing and filtering methods. In the experiments, we leveraged the Rashomon effect creating Rashomon sets on each dataset. We focus on the following research questions not only to examine the impact of balancing methods on models but also the potential mitigating strategies for reducing the predictive multiplicity in the data-centric AI view: (RQ1) How do balancing methods affect the predictive multiplicity of the models in the Rashomon set? (RQ2) What is the effect of filtering methods on reducing predictive multiplicity? (RQ3) Does the impact of filtering methods on predictive multiplicity change based on the complexity of the data? (RQ4) Does the impact of balancing methods on the predictive multiplicity change based on the complexity of the data? (RQ5) How do filtering methods affect the trade-off between predictive multiplicity and model performance? and (RQ6) Can complexity metrics be an indicator of predictive multiplicity? The rest of this paper is structured as follows. Sect. 2 covers the essentials of the Rashomon effect and predictive multiplicity, Sect. 3 outlines the experiments; Sect. 4 interprets the findings, and the final section offers the conclusions.

\section{Related Work}
\subsection{Data-centric vs. Model-centric AI}

The growing interest in data-centric AI reflects a shift in focus from model-centric advancements to improving data quality and reliability. Unlike the traditional model-centric paradigm, which prioritizes model design and hyperparameter tuning while keeping datasets largely static \cite{Zha_et_al_2023_2,Tanov_2023}, data-centric AI emphasizes refining and augmenting data to optimize performance \cite{Jakubik_et_al_2024}. This includes addressing issues like mislabeled, ambiguous, or irrelevant instances \cite{Motamedi_et_al_2021} and leveraging synthetic data to enhance predictive tasks when real data is unavailable \cite{Wang_et_al_2023}. Key functions in this paradigm encompass training data development, inference data preparation, and ongoing data maintenance \cite{Zha_et_al_2023}. The approach has particular significance for small or imbalanced datasets, where dataset optimization is critical for effective AI training \cite{Tanov_2023,Motamedi_et_al_2021}. Ultimately, data-centric AI reframes data not as static input but as a dynamic, evolving asset central to the AI lifecycle \cite{Kumar_et_al_2024}.

\subsection{Balancing methods} 

The problem of class imbalance poses a significant challenge in classification tasks as it degrades model performance \cite{Ortega_et_al_2024}. It has been tackled with various proposed methods, which can be categorized into two groups: algorithmic-level methods \cite{Gu_et_al_2022,Li_et_al_2022}, focused on improving algorithms, and data-level methods \cite{Chawla_et_al_2002,He_et_al_2008}, which involve altering the original dataset to achieve balance called balancing methods-oversampling, undersampling, and several SMOTE-variants are the most commonly used solutions for this problem. Even though there are numerous balancing methods proposed in the literature, none of them is universally superior \cite{Moniz_et_al_2021}. This situation not only burdens researchers to choose the appropriate method but also leaves them vulnerable to the negative effects of these methods on the model. They lead to undesirable shifts in model parameter estimates \cite{Stando_et_al_2024}, alterations in correlations between variables within the model \cite{Patil_et_al_2020}, changes in variable importance \cite{Alarab_and_Prakoonwit_2022}, degradation of model calibration \cite{Goorbergh_et_al_2022}, impacts on model fairness \cite{Simson_et_al_2023}, and the rise of predictive multiplicity driven by the amplification of the Rashomon effect \cite{Cavus_and_Biecek_2024}. Predictive multiplicity can lead to challenges in model selection and decision-making reliability, while also causing issues related to fairness, consistency, ethical accountability, uncertainty, and generalizability \cite{Hsu_and_Calmon_2022,Watson_et_al_2023}. These findings underlined that balancing methods do not work as well as they are thought to and that they should be used responsibly.

\subsection{Filtering methods} 

Variable selection is the critical step in ML, particularly when dealing with noisy or high-dimensional data. Filtering methods are among the most popular techniques for feature selection due to their efficiency and simplicity \cite{Ferreira_et_al_2012,Chandrashekar_et_al_2014,Bommert_et_al_2020}. Two more groups of variable selection methods are called wrapper and embedded methods. Wrapper methods, such as Boruta \cite{Kursa_et_al_2010} and Recursive Feature Elimination \cite{Guyon_et_al_2002}, consider the subset of the variables, and a surrogate ML model is trained. The subgroup is assessed using a performance measure. Thus, it has a higher computational cost than filtering methods. Embedded methods like Lasso, integrate the variable selection step into the modeling process. They balance the computational efficiency of filtering methods and the utility of wrapper methods. The performance of filtering methods is typically evaluated based on their ability to reduce dimensionality while maintaining or improving the predictive accuracy of an ML model. The relevant literature has shown that no single filtering method consistently outperforms others across all datasets \cite{Zheng_et_al_2018}. Instead, the effectiveness of a technique can vary depending on the specific characteristics of the dataset \cite{Solorio_et_al_2020}. For instance, a comprehensive evaluation of 22 filtering methods on 16 high-dimensional classification datasets revealed that while some methods performed well on many datasets, there was no universally superior method \cite{Bommert_et_al_2020}. The correlation techniques are compared with the significance-based techniques and the Spearman correlation coefficient gave the best results improving the accuracy by $92\%$ \cite{Jebadurai_et_al_2022}. Selecting features based solely on statistical significance can lead to overestimated effect sizes. Significant results are more likely to be reported, while non-significant results are often ignored, leading to a biased selection \cite{Bakdash_et_al_2022}.

\subsection{Data complexity} 

Data complexity refers to the difficulty level involved in accurately representing a dataset. Traditional metrics like the imbalance ratio (IR) have been shown to correlate poorly with classifier performance, underscoring the need for more accurate complexity measures to guide preprocessing and model selection \cite{Gottcke_2023}. Imbalance exacerbates the impact of other data characteristics, such as class overlap and challenging decision boundaries, which can undermine model performance \cite{Barella_2021}. Techniques like oversampling and cost-sensitive learning interact with data complexity in distinct ways, with their efficacy varying based on regions of the complexity space and the separability of minority and majority classes \cite{Luengo_et_al_2011,Junior_and_Pisani_2022}. Furthermore, complexity-aware strategies offer opportunities to refine preprocessing steps and select algorithms tailored to imbalanced problems, helping practitioners navigate potential pitfalls \cite{Gottcke_2023}. 

In addition to the above studies, we will examine the effect of preprocessing steps-balancing and filtering methods on the prediction multiplicity leveraging the Rashomon effect. For this purpose, we will conduct experiments on 21 real-world datasets and reach generalizable results.
\subsection{Rashomon effect and predictive multiplicity} 

The Rashomon effect highlights multiple equally good models for an ML task, each achieving near-optimal performance but differing significantly in properties such as fairness, interpretability, and generalization capabilities \cite{Rudin_et_al_2022}, called predictive multiplicity \cite{Marx_et_al_2020}. This diversity offers an opportunity to prioritize domain-specific objectives, such as monotonicity or stability, while avoiding reliance on a single model, which may encode undesired biases or poor generalization \cite{Damour_2021,Semenova_et_al_2022}. Predictive multiplicity within the Rashomon set, often resulting from noisy or underspecified datasets, poses challenges for decision-making, as it can lead to arbitrary or unjustified outcomes \cite{Watson_Daniels_et_al_2024}. Understanding and leveraging this set enables the alignment of ML solutions with ethical, practical, and interpretative requirements, providing actionable insights beyond the conventional focus on accuracy alone \cite{Ciaperoni_et_al_2024,Ganesh_et_al_2024}. This paradigm shift underscores the need for frameworks that quantify and address multiplicity, enhancing the trust and reliability of ML applications in critical domains \cite{Black_et_al_2022,Rudin_et_al_2024}.

\section{Methods}

This section gives the concept of the Rashomon effect, predictive multiplicity, disagreement metrics, and balancing and filtering methods in a general context.

\subsection{Rashomon Effect}

Let $\mathbf{D} = \{ \textbf{x}_i, y_i \}_{i=1}^n$ be a dataset with $n$ observations from $p$ variables, where $\textbf{x}_i = [x_{i1}, x_{i2}, \ldots, x_{ip}] \in \mathbf{X}$ and $y_i \in Y$ is the response vector. Define $F = \{f \mid f \colon \mathbf{X} \to Y\}$ as the space of all predictive models, known as the \textbf{hypothesis space}. Furthermore, let $L \colon F \to \mathbb{R}$ denote the model's loss function. The \textbf{reference model} in the hypothesis space is the model that minimizes the expected value of the loss function:

\begin{equation}
f_R = \operatorname*{argmin}_{f \in F} \mathbb{E}[L(f)].
\end{equation}

\noindent The expected loss $\mathbb{E}[L]$ is approximated by the empirical loss based on the data $\mathbf{D}$. For a given loss function $L$, reference model $f_R$, and a \textbf{Rashomon parameter} $\varepsilon > 0$, the \textbf{Rashomon set} $R_{L,\varepsilon}(f_R)$ is defined as:

\begin{equation}
R_{L,\varepsilon}(f_R) = \{f \in F \mid \mathbb{E}[L(f)] \leq \mathbb{E}[L(f_R)] + \varepsilon\}.
\end{equation}

\noindent Since accessing all possible models in $F$ is infeasible, we focus on an empirical hypothesis space $\hat{F} = \{(f_1, f_2, ..., f_m) \mid f \colon \mathbf{X} \to Y\} \subset F$. Let the model with the minimum empirical loss found in the empirical hypothesis space $\hat{F}$ be the \textbf{empirical reference model}, denoted as $\hat{f}_R$ and formally given as:

\begin{equation}
\hat{f}_R = \operatorname*{argmin}_{f \in \hat{F}} \frac{1}{m} \sum_{i=1}^m L(f(\textbf{x}_i), y_i).
\end{equation}

\noindent For a given loss function $L$ and an empirical reference model $\hat{f}_R$, the \textbf{empirical Rashomon set} is defined as:

\begin{equation}
\hat{R}_{L,\varepsilon}(\hat{f}_R) = \{f \in \hat{F} \mid \frac{1}{m} \sum_{i=1}^m L(f(\textbf{x}_i), y_i) \leq \frac{1}{m} \sum_{i=1}^m L(\hat{f}_R, y_i) + \varepsilon\}.
\end{equation}

\noindent Here, the expected loss $\mathbb{E}[L(f)]$ is approximated by the empirical loss, which is the average of the loss function evaluated over a finite set of observations. This approximation is crucial because calculating the true expected loss is not practical. Thus, the Rashomon effect indicates the existence of similar high-performing models on the same dataset.\\

\noindent \textbf{Predictive multiplicity.} occurs when there exists a model $f \in \hat{R}_{L,\varepsilon}$ such that $f(\textbf{x}_i) \neq f_R(\textbf{x}_i)$ for some $\textbf{x}_i \in \mathbf{X}$ \cite{Marx_et_al_2020}. It illustrates multiple models within the Rashomon set, where these models make conflicting predictions for certain observations while still demonstrating similar prediction performance. The severity of predictive multiplicity depends on the number of conflicting predictions in the dataset. It can be measured from numerous points of view using several metrics introduced \cite{Marx_et_al_2020,Black_et_al_2022,Cavus_and_Biecek_2024,Gomez_et_al_2024,Hamman_et_al_2024,Hsu_et_al_2024}, and we used discrepancy and obscurity because of their comprehensivity on measuring the predictive multiplicity in the level of observations and models.\\

\paragraph{\textbf{Discrepancy}}
The discrepancy $\delta_{\varepsilon}(\hat{f}_R)$ measures the maximum fraction of conflicting predictions between the reference model $f_R$ and the other models in the Rashomon set \cite{Marx_et_al_2020}:

\begin{equation}
\delta_{\varepsilon}(\hat{f}_R) = \max_{f \in \hat{R}_{L,\varepsilon}(\hat{f}_R)} \frac{1}{n} \sum_{i=1}^{n} \mathbbm{1}[f(\textbf{x}_i) \neq \hat{f}_R(\textbf{x}_i)].
\end{equation}

\paragraph{\textbf{Obscurity}}
The obscurity $\gamma_{\varepsilon}(\hat{f}_R)$ calculates the average ratio of conflicting predictions for each observation between the reference model $f_R$ and the other models in the Rashomon set \cite{Cavus_and_Biecek_2024}:

\begin{equation}
\gamma_{\varepsilon}(\hat{f}_R) = \frac{1}{n} \sum_{i=1}^{n} \frac{1}{|\hat{R}_{L,\varepsilon}(\hat{f}_R)|} \sum_{f \in \hat{R}_{L,\varepsilon}(\hat{f}_R)} \mathbbm{1}[f(\textbf{x}_i) \neq \hat{f}_R(\textbf{x}_i)].
\end{equation}

\begin{figure}[ht]
    \centering     
    \small
    \includegraphics[scale=0.29]{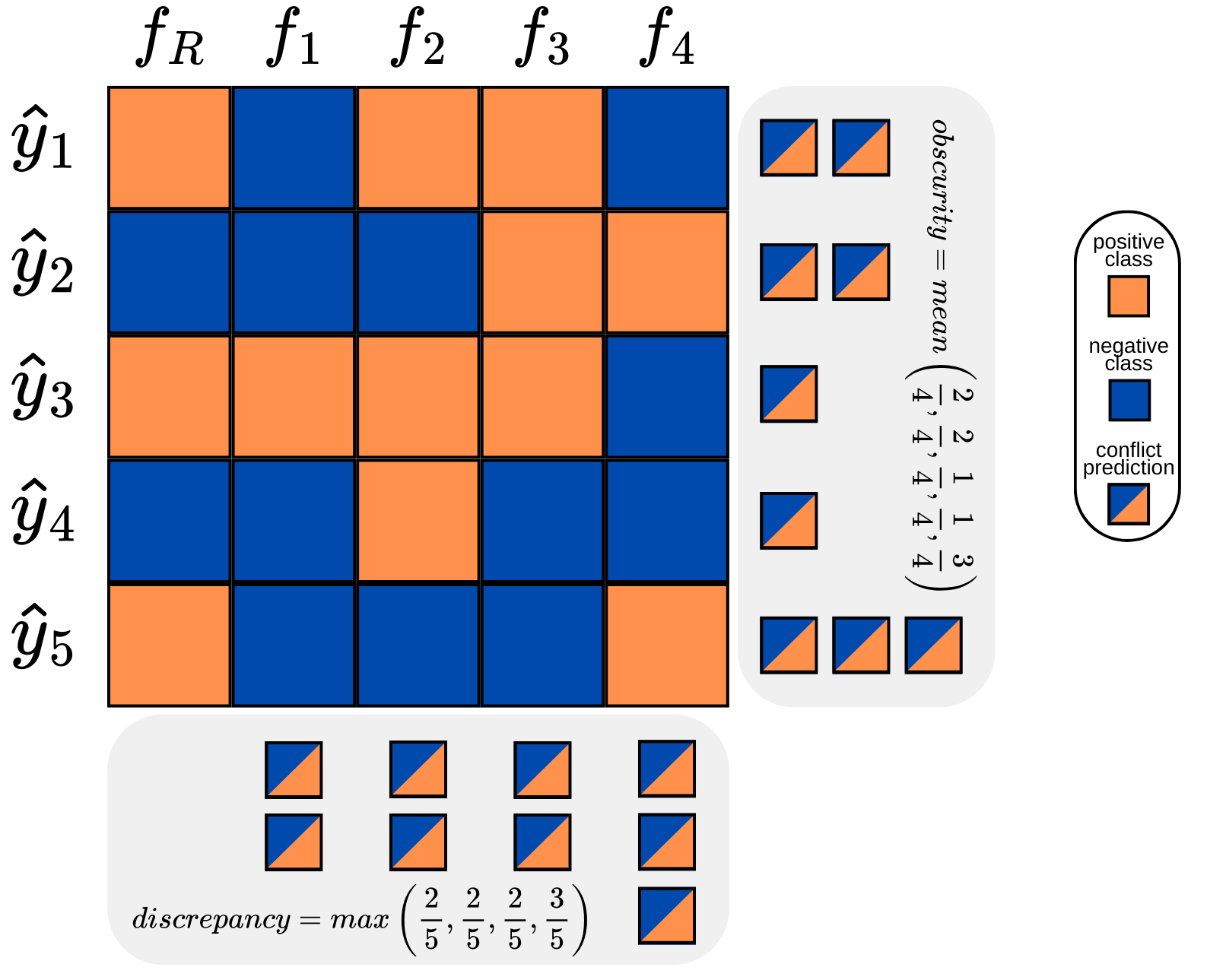}
    \caption{Illustration of a Rashomon cube with size 5 means that comprises five models and observations. The discrepancy is the maximum conflict ratio between the models, and obscurity shows the mean conflict ratio across the observations.}
    \label{fig:rashomon}
\end{figure}

The discrepancy and obscurity offer a granular view of predictive multiplicity by quantifying the variability and conflicts in predictions within the Rashomon set. Discrepancy captures the extent of disagreement across models, while obscurity highlights the degree of uncertainty at the observation level. Figure~\ref{fig:rashomon} illustrates how these metrics are calculated across a Rashomon cube with size $5$, comprising five models and observations. It consists of $1$ reference and $4$ models. The predictions of the reference model, denoted by $\hat{y}_i = \hat{f}_R(\textbf{x}_i)$, are displayed in the first column, with the next columns showing predictions from four different models ($f_1, f_2, f_3, f_4$) within the Rashomon set. The discrepancy signifies the highest conflict ratio between the reference and other models, while obscurity refers to the average conflict ratio across all observations. Determining these metrics requires access to the models within the Rashomon set, a task that poses significant computational challenges \cite{Hsu_et_al_2024,Donnelly_et_al_2024}. 

These metrics provide critical insights into the stability and diversity of model behaviors, making them essential tools for evaluating the impact of pre-processing steps on model selection. To bridge the gap between theoretical model behavior and practical implementation challenges by focusing on these metrics.

\subsection{Balancing Methods}

This section discusses common balancing methods such as Oversampling, Undersampling, Near miss, and SMOTE-based used to address class imbalance in datasets. 

\subsubsection{Oversampling}  
Oversampling increases the representation of the minority class by duplicating its samples.

\begin{equation}
\mathbf{D}' = \mathbf{D} \cup \{(\textbf{x}_i, y_i) \mid y_i \in Y_{\text{min}}\}
\end{equation}

\noindent where $Y_{\text{min}}$ denotes the minority class.
\subsubsection{Undersampling}  
In undersampling, random examples from the majority class are removed:

\begin{equation}
\mathbf{D}' = \mathbf{D} \setminus \{(x_j, y_j) \mid y_j \in Y_{\text{maj}}\}
\end{equation}

\noindent where $Y_{\text{maj}}$ denotes the majority class. Although this method improves class balance, it can lead to information loss.
\subsubsection{Near Miss}  
Near Miss selects samples from the majority class closest to the minority class \cite{Mani_Zhang_2003}. The distance between two samples $x_i$ and $x_j$ is given by:

\begin{equation}
d(x_i, x_j) = \sqrt{\sum_{k=1}^{n} (x_{ik} - x_{jk})^2}.
\end{equation}
\subsubsection{Synthetic Minority Oversampling Technique}  
Synthetic Minority Oversampling Technique (SMOTE) generates synthetic samples by interpolating between a minority sample $x_i$ and one of its neighbors $x_j$ \cite{Chawla_et_al_2002}:

\begin{equation}
x_{\text{new}} = x_i + \lambda \cdot (x_j - x_i), 
\end{equation}

\noindent where $\lambda \in [0, 1]$ and  $x_{\text{new}}$ denotes the synthetic sample and $\lambda$ controls the interpolation between the observations $x_i$ and $x_j$.
\subsubsection{Adaptive Synthetic Sampling Approach for Imbalanced Learning}  
Adaptive Synthetic Sampling Approach Imbalanced Learning (ADASYN) generates more synthetic samples in regions with low density of minority classes \cite{He_et_al_2008}. The local density difference $r_i$ for a minority sample $x_i$ is defined as

\begin{equation}
r_i = \frac{\Delta_i}{k},
\end{equation}

\noindent where $ \Delta_i $ is the number of neighbors in the majority class among the neighbors in $k$-nearest. The number of synthetic samples for each minority instance is:

\begin{equation}
G_i = \text{round}(r_i \cdot G),
\end{equation}

\noindent where $G$ is the number of synthetic samples. The new sample is the following.

\begin{equation}
x_{\text{new}} = x_i + \lambda \cdot (x_j - x_i),
\end{equation}

\noindent where $\lambda \in [0, 1]$.
\subsubsection{Borderline Synthetic Minority Oversampling Technique}  
Borderline Synthetic Minority Oversampling Technique (BLSMOTE) generates synthetic samples for minority instances near the decision boundary \cite{Han_et_al_2005}. For a minority instance $x_i$, the synthetic sample is:

\begin{equation}
x_{\text{new}} = x_i + \lambda \cdot (x_j - x_i),
\end{equation}

\noindent where $\lambda \in [0, 1]$.
\subsubsection{Density-Based SMOTE}  
Density-based SMOTE (DBSMOTE) adjusts the number of synthetic samples based on local density \cite{Bunkhumpornpat_et_al_2012}:

\begin{equation}
G_i = \frac{1}{d_i} \cdot G,
\end{equation}

\noindent where $d_i$ is the local density of the minority instance $x_i$.
\subsubsection{Relocating Safe-Level SMOTE}  
Relocating Safe-Level SMOTE (RSLSMOTE) generates synthetic samples only for safe minority instances \cite{Siriseriwan_Sinapiromsaran_2016}. The safe level $s(x_i)$ is defined as:

\begin{equation}
s(x_i) = \frac{|\{x_j \in \mathcal{N}(x_i) \mid y_j = y_i\}|}{|\mathcal{N}(x_i)|},
\end{equation}

\noindent where $ \mathcal{N}(x_i) $ is the neighbourhood of $x_i$. RSLSMOTE relocates unsafe instances to safer regions before generating synthetic samples.
\subsubsection{Adaptive Neighbor SMOTE}  
Adaptive Neighbor SMOTE (ANSMOTE) adapts the placement of synthetic samples based on local neighborhood structures \cite{Siriseriwan_Sinapiromsaran_2017}. The synthetic sample is:

\begin{equation}
x_{\text{new}} = x_i + \lambda \cdot (x_j - x_i),
\end{equation}

\noindent where $\lambda \sim \mathcal{U}(0, 1)$ and $ \mathcal{U}(0, 1) $ denotes a uniform distribution over the interval $[0, 1]$.
\subsubsection{Safe-Level SMOTE}  
Safe-Level SMOTE (SLSMOTE) ensures that synthetic samples are generated from minority instances that are considered safe \cite{Bunkhumpornpat_et_al_2009}. The safe level $s(x_i)$ of a minority instance $x_i$ is defined as:

\begin{equation}
s(x_i) = \frac{|\{x_j \in \mathcal{N}(x_i) \mid y_j = y_i\}|}{|\mathcal{N}(x_i)|},
\end{equation}

\noindent where $ \mathcal{N}(x_i) $ denotes the $k$-nearest neighbors of $x_i$. If $s(x_i)$ is sufficiently high, a new synthetic sample is generated as follows:

\begin{equation} 
x_{\text{new}} = x_i + \lambda \cdot (x_j - x_i),
\end{equation}

\noindent where $\lambda \in [0, 1]$. This approach avoids the generation of synthetic samples near the decision boundary, ensuring better class separation.
\subsection{Filtering Methods}
In this section, the filtering methods are Correlation and Significance Tests are given. 

\subsubsection{Correlation Testing}

The correlation analysis aims to quantify the relationship between the variables $\mathbf{x}_i$ and the target variable $y_i$. The correlation coefficient $r_{xy}$ between a variable $x_j$ and the target $y$ is calculated as:

\begin{equation}
    r_{xy} = \frac{\text{Cov}(x_j, y)}{\sigma_{x_j} \sigma_y}
\end{equation}

\noindent where $\text{Cov}(x_j, y)$ is the covariance between the variable $x_j$ and $y$, and $\sigma_{x_j}$ and $\sigma_y$ are the standard deviations of $x_j$ and $y$, respectively. For the hypothesis test of correlation, the following hypotheses are tested:

\begin{equation}
    H_0: r_{xy} = 0 \quad \text{vs} \quad H_1: r_{xy} \neq 0
\end{equation}

\noindent If $p_{r} < \alpha$, the null hypothesis is rejected, indicating a significant correlation. Then, the Benjamini-Hochberg (BH) procedure \cite{Benjamini_and_Hochberg_1995} is applied to adjust the p-values:

\begin{equation}
    p_{r,\text{adj}} = \min\left(1, \frac{p_r \cdot m}{k}\right),
\end{equation}

\noindent where $m$ is the total number of tests and $k$, is the rank of the p-value. The BH procedure allows us to control the false discovery rate when multiple hypotheses are tested. Adjusting the significance level based on the rank of the p-values ensures that the expected proportion of incorrectly rejected null hypotheses remains below a specified threshold. Thus, the set of significant variables $S_{\text{cor}}$ is defined as follows:

\begin{equation}
    S_{\text{cor}} = \{x_j \in \mathbf{X} \mid p_{r,\text{adj}} < \alpha\}
\end{equation}

\subsubsection{Significance Testing}

The Wilcoxon rank sum test \cite{Wilcoxon_1992} is performed to assess differences in variable distributions between the two classes of our target variable. It assesses whether two independent samples (\(x_j\) for \(y=0\) and \(y=1\)) come from the same distribution. The null hypothesis defines the test.

\begin{equation}
H_0: F_0(x) = F_1(x) \quad \text{vs} \quad H_1: F_0(x) \neq F_1(x),
\end{equation}

\noindent where \(F_0\) and \(F_1\) are the cumulative distribution functions of the two classes. The p-value \(p_{\text{sig}}\) is calculated for each variable \(x_j\) and the BH method is used:

\begin{equation}
p_{\text{sig, adj}} = \min\left(1, \frac{p_{\text{sig}} \cdot m}{k}\right).
\end{equation}

\noindent The set of significant variables \(S_{\text{sig}}\) is defined as:

\begin{equation}
S_{\text{sig}} = \{x_j \in \mathbf{X} \mid p_{\text{sig, adj}} < \alpha\}.
\end{equation}

\noindent The final set of selected variables \(S\) is then defined as

\begin{equation}
S = S_{\text{cor}} \cap S_{\text{sig}}.
\end{equation}

\noindent In this study, these predictive modeling frameworks are utilized to select variables, denoted by the hypothesis space \(F\).

\section{Experiments}
In this section, we perform experiments to examine the impact of balancing techniques on the Rashomon effect in terms of predictive multiplicity and the reducing effect of processing methods, regarding the complexity of data. Ten balancing techniques are explored: Oversampling, SMOTE, Undersampling, Near miss, ADASYN, ANSMOTE, BLSMOTE, DBSMOTE, RSLSMOTE, and SLSMOTE. During the balancing phase, adjustments are made to the training set, leaving the test set unchanged. The resampling ratio (the imbalance ratio post-balancing) is set to $1$, which means the perfect balance between the classes. The filtering methods Correlation test and Significance test are used to select the relevant variables.

We leveraged the Rashomon perspective to examine the impact of processing methods on predictive multiplicity. It works on a model set that consists of models that exhibit similar high prediction performance. There are several ways such as using different seeds in data splitting, various hyperparameter settings, different model families, pre-trained models, or any combination of these to create the Rashomon set. We utilize a tree-based AutoML tool \texttt{forester} \cite{forester} which combines different models and hyperparameter settings because it is easy to use. It also provides the advantage of controlling the Rashomon set size using parameters from the tool's Bayesian optimization component. The Rashomon parameter $\epsilon$ is assigned a value of $0.05$. Furthermore, the number of optimization rounds (\texttt{bayes\_iter}) is defined as $5$, and the number of models trained (\texttt{random\_ evals}) within the \texttt{forester} training function is fixed at $10$. The Rashomon set is derived from tasks within an imbalanced benchmark dataset, the details of which are given in the following section. 
\subsection{Dataset}
\label{dataset}

We used imbalanced benchmark data sets \cite{Stando_et_al_2024} to analyze the impact of data-centric AI solutions on imbalanced learning and predictive multiplicity. It consists of several data sets of various sizes with at least $1000$ observations and imbalance ratios higher than $1.5$ for the binary classification task; the details of the datasets are given in Table~\ref{tab:dataset}. 

\begin{table}[H]
\centering     
\scriptsize
\caption{The characteristics of the 21 datasets used in the empirical analysis. IR stands for the imbalance ratio, and Cluster is defined as the complexity clusters of the data to which they belong.}
\begin{tabular}{lrrrr}\toprule
Dataset                       & IR       & \#Samples & \#Variables & Cluster\\ \midrule
\texttt{spambase}             & 1.54     & 1055      & 42          & 1\\
\texttt{MagicTelescope}       & 1.84     & 19020     & 11          & 2\\
\texttt{steel\_plates\_fault} & 1.88     & 8378      & 121         & 3\\
\texttt{qsar-biodeg}          & 1.96     & 1055      & 17          & 1\\
\texttt{phoneme}              & 2.41     & 5404      & 6           & 2\\
\texttt{jm1}                  & 4.17     & 10885     & 22          & 1\\
\texttt{SpeedDating}          & 4.63     & 4601      & 58          & 1\\
\texttt{kc1}                  & 5.47     & 2109      & 22          & 1\\
\texttt{churn}                & 6.07     & 5000      & 21          & 1\\  
\texttt{pc4}                  & 7.19     & 1458      & 38          & 1\\
\texttt{pc3}                  & 8.77     & 1563      & 38          & 1\\
\texttt{abalone}              & 9.68     & 4177      & 7           & 1\\
\texttt{us\_crime}            & 12.29    & 1994      & 101         & 3\\    
\texttt{yeast\_ml8}           & 12.58    & 2417      & 104         & 3\\
\texttt{pc1}                  & 13.40    & 1109      & 22          & 1\\
\texttt{ozone\_level\_8hr}    & 14.84    & 2534      & 73          & 3\\
\texttt{wilt}                 & 17.54    & 4839      & 36          & 3\\
\texttt{wine\_quality}        & 25.77    & 4898      & 12          & 1\\
\texttt{yeast\_me2}           & 28.10    & 1484      & 9           & 2\\
\texttt{mammography}          & 42.01    & 11183     & 7           & 2\\
\texttt{abalone\_19}          & 129.53   & 4177      & 7           & 3\\\bottomrule        
\end{tabular}
\label{tab:dataset}
\end{table}

\begin{figure}
    \centering     
    \includegraphics[width=0.8\linewidth]{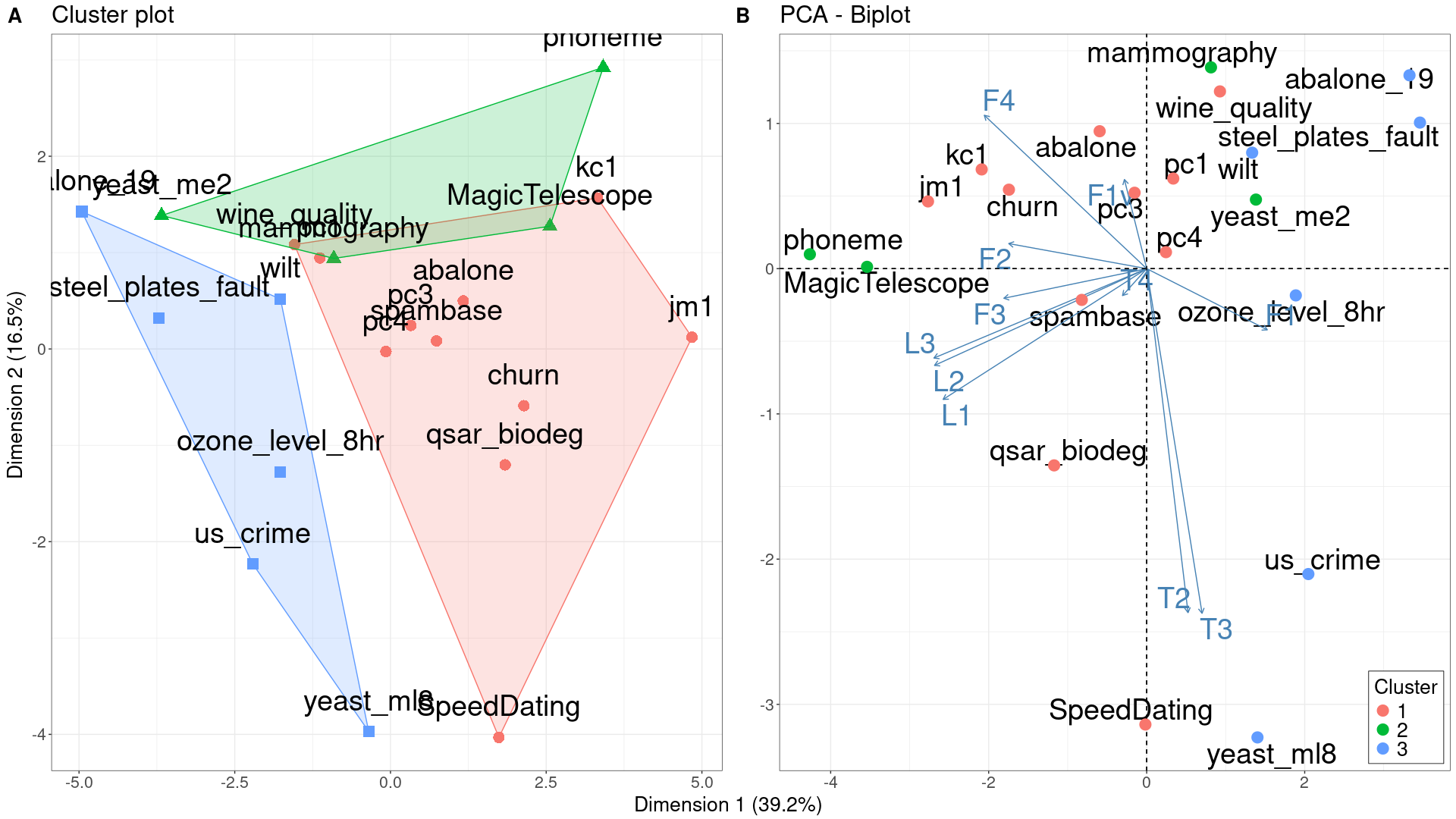}
    \caption{The similarity analysis of datasets from our benchmark. Similarity is presented with the cluster plot and the PCA plot. Similarity is calculated based on the data complexity metrics}
    \label{fig:dataset_plot}
\end{figure}

The class imbalance ratio is not always a useful indicator of the difficulty of imbalanced learning problems. Therefore, data sets are clustered according to their complexity, and statistics are used to measure the complexity of imbalanced data sets. The explanations and the values of the complexity metrics for each data are given in Appendix~\ref{sec:appendix}. We follow this path because data complexity is measured in different dimensions and complexity cannot be fully reflected with a single metric. The k-means algorithm \cite{MacQueen_1967} is used for clustering and the results are shown in Figure~\ref{fig:dataset_plot}.

The Cluster Plot classifies the data sets into clusters, revealing shared variables within each group. Grouping data sets around specific clusters suggests similarities in data types or problem domains. The PCA-Biplot provides an alternative perspective of the same data, highlighting the influence of variables on each dataset and explaining variance across datasets while minimizing information loss through dimensionality reduction. These findings demonstrate that data sets can be effectively clustered into three groups and presented based on their characteristics. The complexities of the data sets are identified according to the clusters to which they belong in the experiments.

\section{Results}
This section involves the results of experiments conducted to investigate the research questions addressed in the paper. The results are calculated using the models of the Rashomon sets created on the datasets. Two types of plots—boxplot and 2d density—are utilized to present the results. The 2d density plots are used to visualize the predictive multiplicity of the Rashomon sets consisting of the nearly-high-accurate models trained on the datasets in terms of the disagreement metrics—discrepancy and obscurity. The triangle icons in the 2d density plots indicate the average values (median) of the disagreement metric corresponding to the axis. Moreover, the statistical hypothesis tests—Kruskal-Wallis test \cite{Kruskal_and_Wallis_1952}, Friedman test \cite{Friedman_1937}, and Dunn pairwise posthoc comparison test \cite{Dunn_1964}—are used to provide more robust findings. 

\subsection*{\textbf{RQ1. How do the balancing methods affect the predictive multiplicity of the models in the Rashomon set?}}

\begin{figure}[ht]
    \centering
    \includegraphics[width=0.8\linewidth]{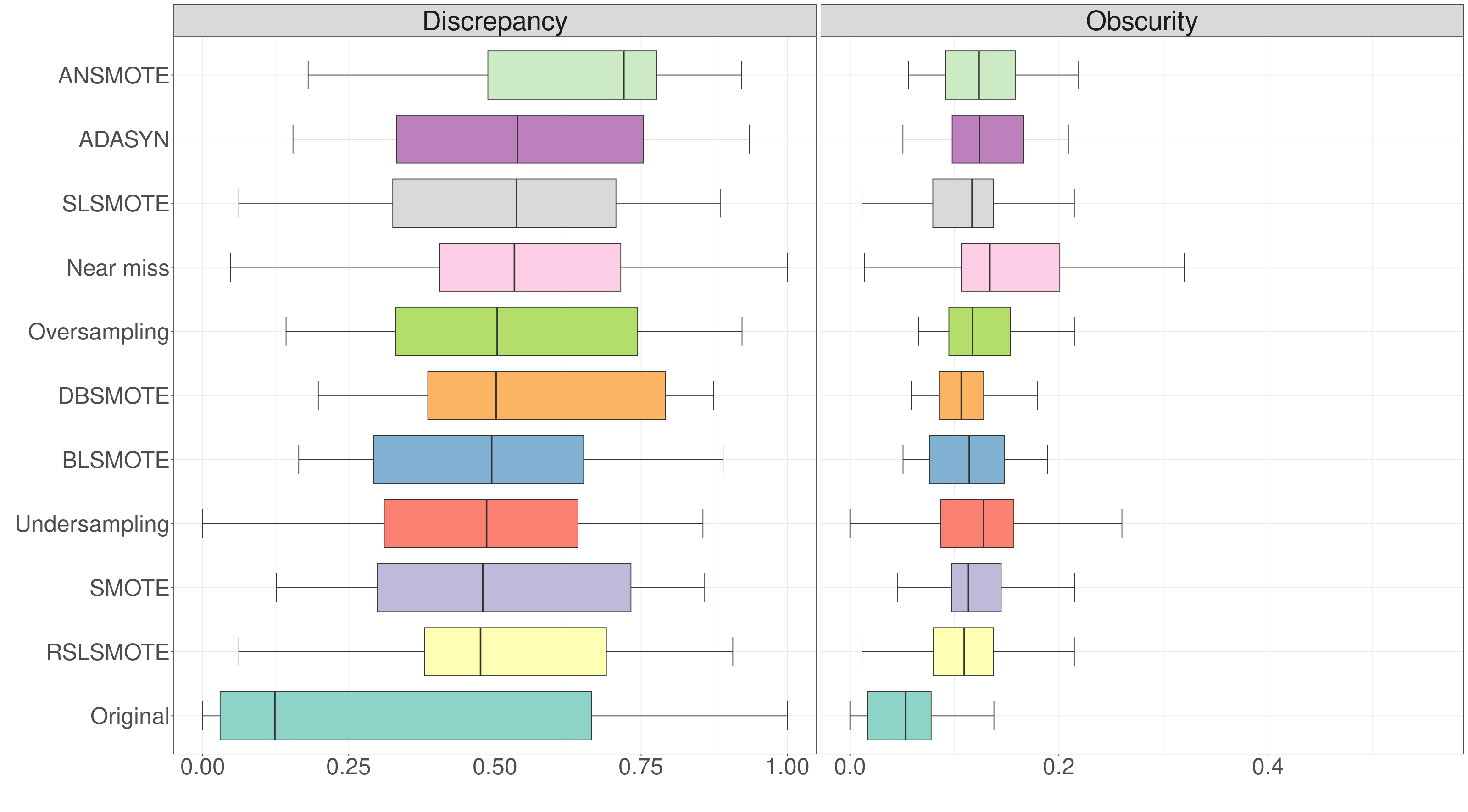}
    \caption{The distribution of the disagreement metrics obscurity and discrepancy for the balancing methods.}
    \label{fig:rash_boxplot}
\end{figure}

Figure~\ref{fig:rash_boxplot} compares the distribution of two disagreement metrics, discrepancy, and obscurity, between various balancing methods, with the Original serving as the reference condition without any balancing applied. Regarding discrepancy, the original dataset exhibits the widest variability, but the lowest average in median compared to the balancing methods. All balancing techniques inflate the average discrepancy, suggesting inconsistency between models. The ANSMOTE method differs negatively from other balancing methods with its highest average discrepancy value. For obscurity, all balancing methods tend to maintain relatively higher values compared to the original dataset, indicating that these methods deteriorate model clarity. Near-miss and undersampling methods generally result in slightly higher obscurity than others, potentially reflecting their impact on reducing data quantity. Overall, the balancing methods appear to inflate both discrepancy and obscurity when compared to the original dataset, with some methods like ANSMOTE, Near miss, and Undersampling demonstrating an increasing impact on predictive multiplicity.

In addition to visual comparisons, statistical tests provide more robust findings. The Kruskal-Wallis test was performed to test the difference of average obscurity and discrepancy of Rashomon sets trained on datasets balanced with different balancing methods and the original (imbalanced) dataset. The KW test results revealed that there were statistically significant differences in the means of discrepancy ($\chi^2_{KW} (10) = 43.98, p < 0.0001$***) and obscurity ($\chi^2_{KW} (10) = 113.30, p < 0.0001$***). When the Dunn post hoc pairwise comparison test was applied to determine which methods differed, it was found that the predictive multiplicity of the Rashomon set associated with the original dataset significantly differed from the predictive multiplicity of Rashomon sets associated with datasets balanced using all other methods.  

\subsection*{\textbf{RQ2. What is the effect of filtering methods on reducing predictive multiplicity?}}

\begin{figure}[H]
    \centering     
    \small
    \includegraphics[width=0.8\linewidth]{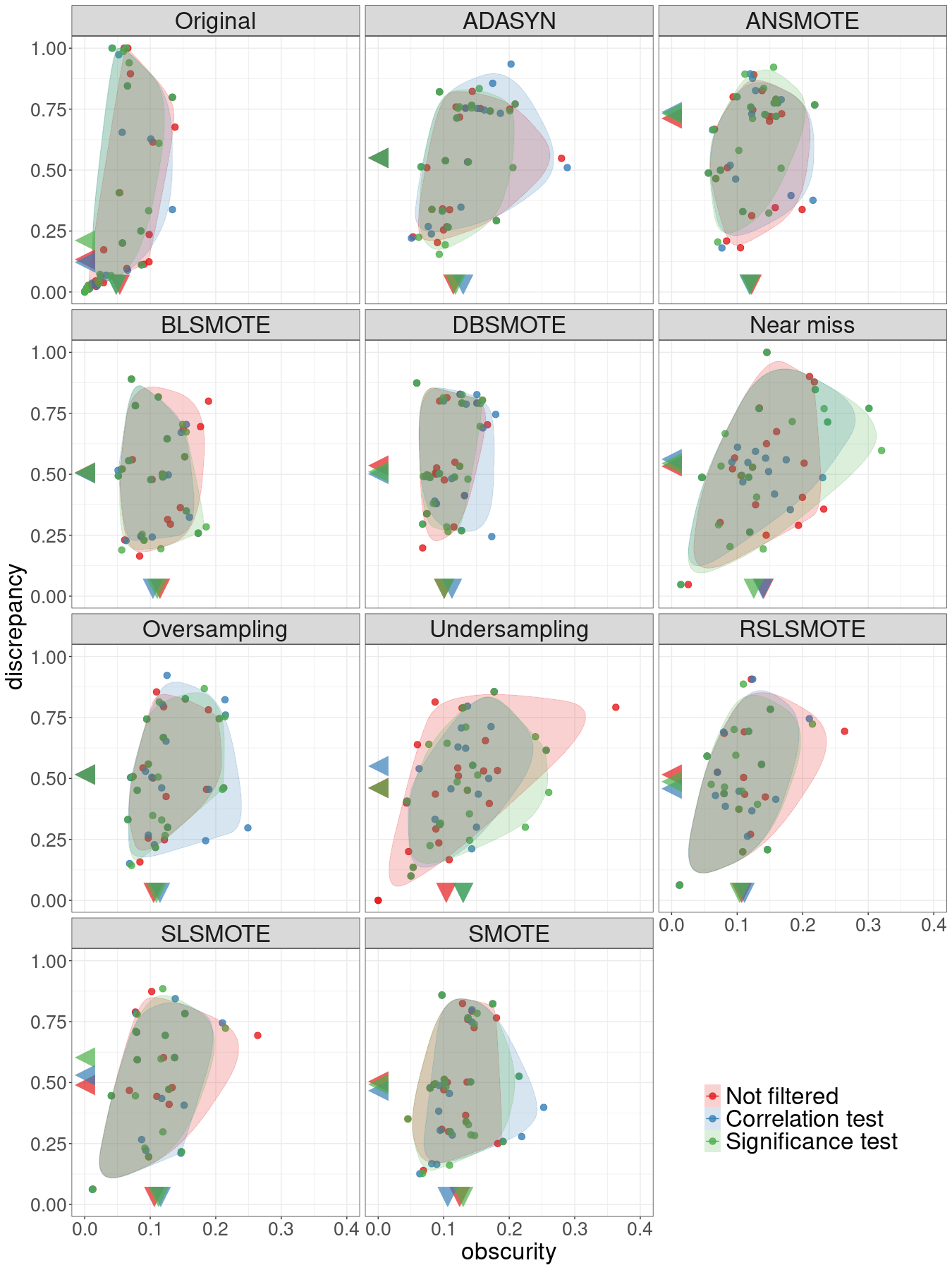}
    \caption{The 2d density plot of the disagreement metrics obscurity and discrepancy for the balancing and filtering methods.}
    \label{fig:rash_for_balancing}
\end{figure}

The density of the disagreement metrics obscurity and discrepancy for different balancing and filtering methods is given as 2d density plots in Figure~\ref{fig:rash_for_balancing}. It shows that the different balancing methods lead the Rashomon set to exhibit varying behaviors in terms of predictive multiplicity under the two filtering approaches. For the original data, without filtering (aka Not Filtered), Significance Test approaches display a wider range of discrepancy values, indicating higher multiplicity in model predictions. Specifically, without filtering, the discrepancy spans a wide range from approximately $0.1$ to $1$, dense around $0.75$, while obscurity extends to $0.3$. The Correlation Test, on the other hand, produces a more focused distribution, with the average discrepancy falling below $0.5$ and the obscurity around $0.1$. Similarly, the Significance Test reduces variability, yielding a compact distribution with an average discrepancy close to $0.4$.

The ADASYN and ANSMOTE reduce the range of discrepancy values, particularly when combined with the filtering methods. ADASYN shows a discrepancy range of $0.2$ to $0.9$, and averages around $0.4$ to $0.5$ under filtering methods, while without filtering, it results in a wider spread with a median near $0.6$. ANSMOTE exhibits similar behavior, with more compact distributions under filtering methods, where the average is around $0.3$ to $0.4$ for the discrepancy and just above $0.1$ for obscurity. Not using any filtering method again results in greater variability.

BLSMOTE, DBSMOTE, and SLSMOTE demonstrate the narrowest discrepancy ranges. DBSMOTE, in particular, achieves the most compact distributions across both metrics, with a discrepancy ranging from approximately $0.2$ to $0.7$ and an obscurity tightly confined between $0.05$ and $0.15$. Its average discrepancy is around $0.3$ under the Significance Test, further enhancing its compactness. BLSMOTE performs similarly but with slightly broader distributions, particularly under the Not Filtered approach, where the average discrepancy rises to $0.6$. SLSMOTE also performs well, with focused distributions under the Significance Test, where the average discrepancy approximates $0.3$.

Other methods such as Oversampling and Near Miss exhibit different behaviors. Oversampling results in broader distributions without filtering and Correlation Test approaches, and averages near $0.6$ for the discrepancy. The Significance Test, however, compresses the data significantly, reducing variability and lowering the average discrepancy to around $0.4$. Near Miss shows wider spreads overall, with an average of around $0.5$ under filtering methods, although it tends to cluster at lower obscurity values.

RLSMOTE and SMOTE demonstrate balanced performances, with discrepancies ranging from approximately $0.3$ to $0.8$ and obscurity ranging from $0.05$ to $0.2$. Although RLSMOTE achieves narrower distributions under the Significance Test, SMOTE also benefits from filtering methods, with average discrepancies concentrated around $0.4$ and obscurity near $0.1$. However, not filtering continues to yield broader distributions for both methods.

In general, filtering methods play a critical role in reducing predictive multiplicity. Not using any filtering consistently results in the widest distributions, reflecting higher variability and inconsistency. The Correlation Test confines data into more compact spaces by removing high-correlation variables, while the Significance Test achieves the most consistent and narrowest distributions across balancing methods. Balancing methods such as DBSMOTE and SLSMOTE are the most effective, particularly when combined with the Significance Test, producing compact and reliable distributions. These findings highlight the importance of selecting appropriate filtering and balancing combinations to optimize model performance and stability of predictions in ML applications.

To test the difference in average obscurity and discrepancy of Rashomon sets trained on datasets after filtering the variables. The results of the KW test revealed that there are no statistically significant differences in the average obscurity ($\chi^2_{KW}(2) = 0.50, p = 0.779$) and the discrepancy ($\chi^2_{KW}(2) = 0.26, p = 0.879$).  

We considered balancing methods as a block effect. Then the Friedman test was conducted to compare the mean obscurity and discrepancy of Rashomon sets created by models after variable selection using filtering methods, and it was found that only the obscurity values show a statistically significant difference ($\chi^2_{F}(2) = 9.25, p = 0.010$*). Consequently, it is observed that the combined effects of filtering methods and balancing methods significantly increase the obscurity values.

\subsection*{\textbf{RQ3. Does the effect of filtering methods on predictive multiplicity change based on the complexity of the data?}}

\begin{figure}[H]
    \centering     
    \small
    \includegraphics[width=1\linewidth]{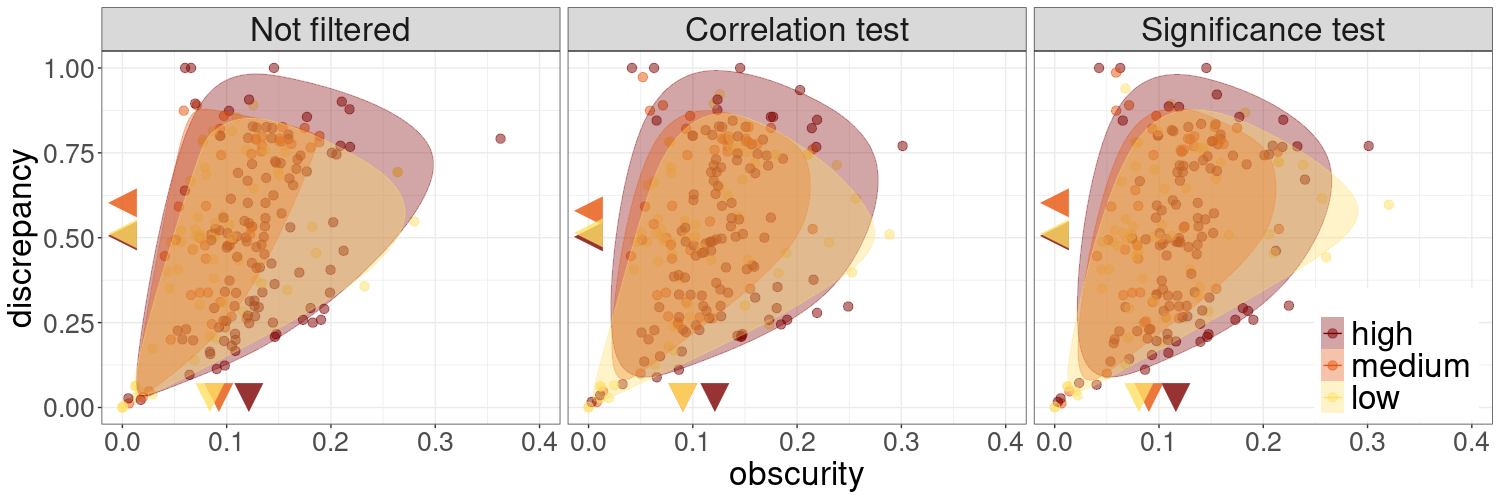}
    \caption{The 2d density plot of the disagreement metrics obscurity and discrepancy for the filtering methods and the complexity of the dataset.}
    \label{fig:rash_for_filter}
\end{figure}

Figure~\ref{fig:rash_for_filter} highlights the relationship between obscurity and the discrepancy under different filtering methods based on the complexity of the dataset. It demonstrates that filtering methods significantly moderate the effects of balancing methods on predictive multiplicity, particularly in datasets with medium to high complexity. For instance, when combined with DBSMOTE or ANSMOTE, the Significance Test leads to compact Rashomon sets with low discrepancy values around $0.3$ and obscurity $\sim 0.1$. In contrast even when filtered, Near Miss and ADASYN, exhibit higher variability, particularly in complex datasets. This underscores the critical role of filtering in mitigating the potential negative effects of balancing methods.

When no filtering is applied, the data exhibit the widest spread, with convex hulls indicating substantial overlap among complexity groups. High-discrepancy and high-obscurity points are concentrated in the upper-right region, while lower values cluster near the origin. The averages without filtering are approximately $0.1$ for obscurity and $0.55$ for discrepancy across all complexity levels, reflecting considerable predictive multiplicity.

Applying the Correlation Test reduces the spread of medium- and high-complexity datasets. It effectively narrows the distributions, particularly for datasets with high complexity. The average values shift slightly, with the obscurity averages remaining near $0.1$ and the discrepancy averages dropping to around $0.5$ for medium and high complexity. The overlap is reduced compared to the without filtering, suggesting the role of filtering methods in limiting variability across Rashomon sets.

The Significance Test further sharpens group boundaries and creates the most compact distributions, particularly for medium and high-complexity datasets. Average obscurity remains consistent at about $0.1$, but discrepancy averages decrease more significantly to approximately $0.45$ for high-complexity datasets. This indicates that the Significance Test is more effective in reducing predictive multiplicity by focusing on statistically significant features, thus narrowing variability and improving model consistency.

To test the difference in mean obscurity and discrepancy of Rashomon sets trained on imbalanced (aka original) datasets according to their complexities. The KW test reveals significant differences in mean obscurity ($\chi^2_{KW}(2) = 8.68, p = 0.0130$*) and discrepancy ($\chi^2_{KW}(2) = 13.68, p = 0.0011$**) between the complexities of the dataset when filtering methods are applied. These differences highlight the varying effects of the complexity of the dataset on the predictive multiplicity metrics. 

In summary, the choice of filtering method impacts predictive multiplicity differently based on the complexity of the dataset. The Significance Test consistently produces more compact and focused distributions, particularly for medium- and high-complexity datasets, making it the most effective at reducing variability. Conversely, the Not Filtered condition leads to broader spreads and higher predictive multiplicity. These findings underscore the importance of selecting appropriate filtering techniques to achieve robust and consistent ML models.

\subsection*{\textbf{RQ4. Does the effect of balancing methods on the predictive multiplicity change based on the complexity of the data?}}

\begin{figure}[H]
    \centering     \small
    \includegraphics[width=0.8\linewidth]{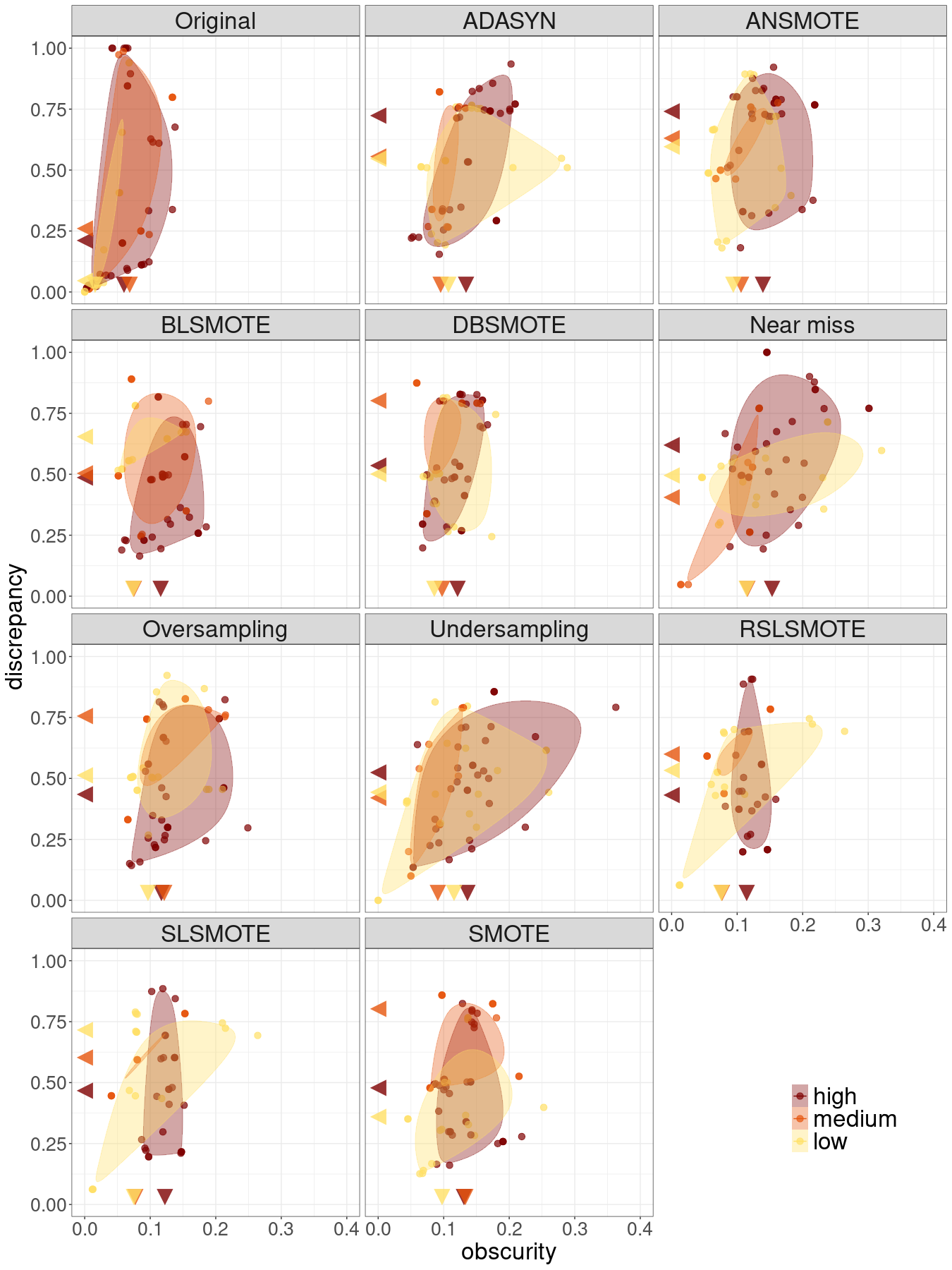}
    \caption{The 2d density plot of the disagreement metrics obscurity and discrepancy for the balancing methods and the complexity of the dataset.}
    \label{fig:rash_for_cluster}
\end{figure}

Figure~\ref{fig:rash_for_cluster} illustrates how balancing methods affect predictive multiplicity, particularly data complexity, measured by obscurity and discrepancy. Models without balancing typically show a wide range of discrepancies, indicating substantial multiplicity in predictions, while obscurity remains relatively low, reflecting limited variation in the predictions. Balancing methods alter these distributions in different ways depending on the complexity of the dataset.

ADASYN generally produces uniform distributions across both metrics. Still, certain regions exhibit slight increases in discrepancy, with discrepancy values remaining constant at around $0.5$ and obscurity near $0.15$ in the medium and high-complexity datasets. This suggests infrequent increases in inconsistency and predictive multiplicity. Similarly, ANSMOTE controls discrepancies relatively well, with discrepancy values around $0.4$ and obscurity slightly above $0.1$. However, it shows a broader range of obscurity, particularly for high-complexity datasets, indicating more variability in model behavior.

BLSMOTE stands out for its compact density region, with low average values for both metrics—approximately $0.3$ for discrepancy and $0.1$ for obscurity—indicating it effectively minimizes predictive multiplicity. DBSMOTE, on the other hand, maintains obscurity values approximately $0.1$. Still, it exhibits discrepancy values near $0.5$, suggesting it balances the trade-off between predictive multiplicity and stability.

In contrast, Near Miss significantly expands the range of both metrics, with obscurity values increasing to around $0.2$ and discrepancy values exceeding $0.6$ for high-complexity datasets. This indicates a higher predictive multiplicity in predictions. Undersampling similarly results in higher discrepancies near $0.6$, but keeps obscurity low around $0.1$, reflecting a significant divergence in model predictions despite low multiplicity.

SMOTE and RSL-SMOTE achieve stable distributions with low obscurity and moderate discrepancies between data complexities. SMOTE consistently maintains obscurity values around $0.1$ and discrepancy values between $0.4$–$0.5$, suggesting stable predictions. RSL-SMOTE performs similarly, with even lower obscurity values $\sim 0.05$ and moderate discrepancies $\sim 0.4$, which make it highly effective in reducing predictive multiplicity while maintaining stability.

Oversampling results in low to medium values around $0.1$–$0.15$ but exhibits a wide range of discrepancies exceeding $0.5$ in more complex datasets. This indicates reduced uncertainty but increased variability in model outcomes. ADASYN and Near Miss tend to exacerbate predictive multiplicity, especially in high-complexity scenarios, whereas RSL-SMOTE and SMOTE remain stable and reliable across different dataset complexities.

In general, the effect of the balancing methods is heavily dependent on the complexity of the data set. RSL-SMOTE and BLSMOTE excel in both metrics, achieving low discrepancy and obscurity values, particularly for simpler datasets. At the same time, Near Miss and ADASYN enhance predictive multiplicity as complexity increases. This underscores the importance of selecting balancing methods that align with the underlying data complexity to ensure robust and reliable ML models.

The Friedman test, which evaluates the combined effects of filtering and balancing methods on obscurity and discrepancy, is conducted. It found no statistically significant differences both in terms of discrepancy ($\chi^2_{F} (2) = 0.66667, p = 0.7165$) and obscurity ($\chi^2_{F} (2) = 0.66665, p = 0.7166$). This suggests that while filtering methods independently influence predictive multiplicity, their combined effects with balancing methods do not further alter the disagreement metrics substantially.

\subsection*{\textbf{RQ5. What is the impact of filtering methods on the trade-off between predictive multiplicity and model performance?}}

\begin{figure}[H]
    \centering
    \includegraphics[width=0.6\linewidth]{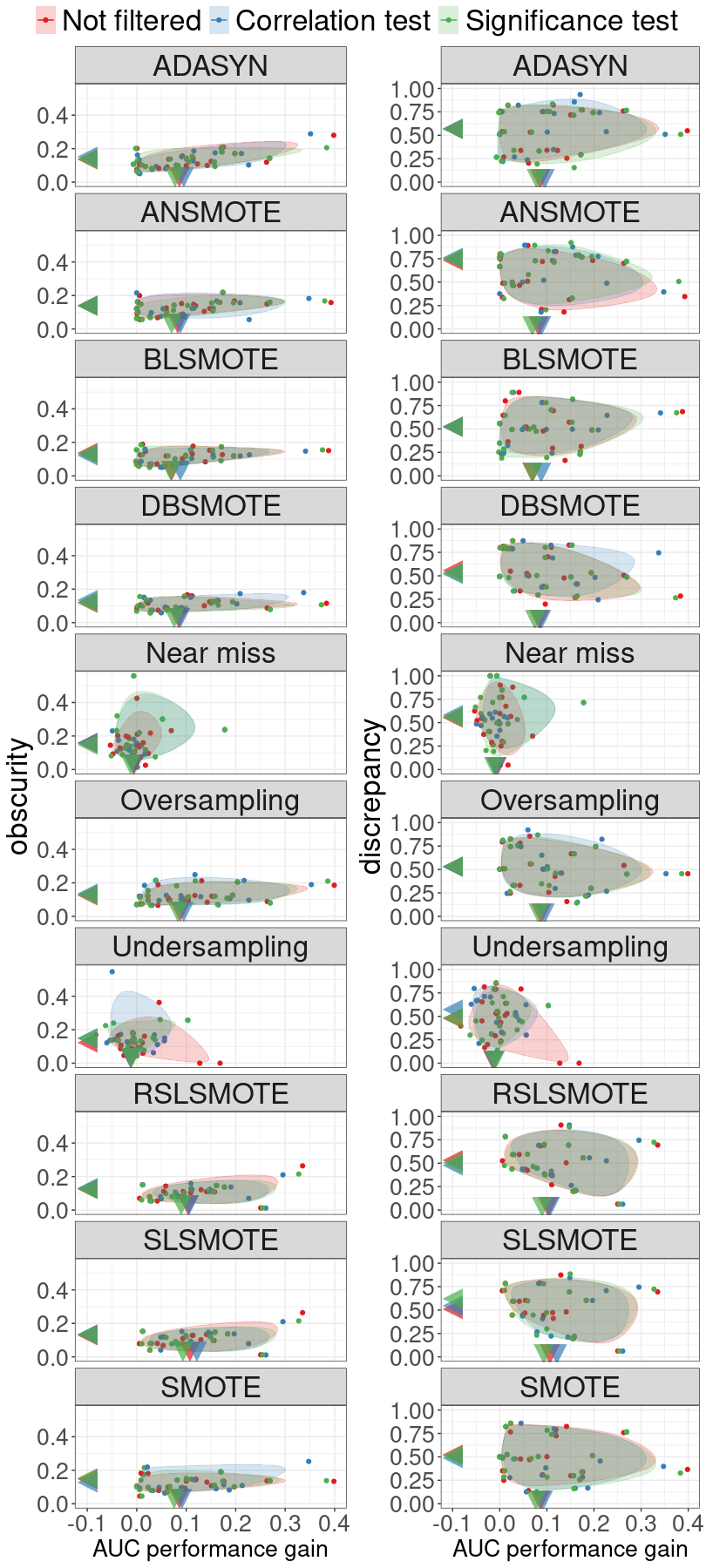}
    \caption{The performance gain plots of obscurity and discrepancy for the balancing and the filtering methods.}
    \label{fig:performance_gain}
\end{figure}

Figure~\ref{fig:performance_gain} shows the trade-off between performance gain, measured as AUC improvement, and disagreement metrics. In particular, for SMOTE, the median AUC performance gain ($0.1$) aligns with moderate increases in obscurity ($0.05$) and discrepancy ($0.5$). Similarly, ADASYN values indicate comparable trends, with an AUC performance gain of approximately $0.09$, average obscurity values near $0.07$, and discrepancy values at $0.48$. This suggests that both methods balance predictive improvement with diversity within the Rashomon set.

The medians are markedly different for undersampling methods such as Near Miss. The performance gain is relatively low ($0.02$), but the average values of both obscurity ($0.2$) and discrepancy ($0.7$) are significantly higher, highlighting a trade-off where increased model disagreement accompanies limited predictive improvement. This pattern is consistent with general undersampling, showing a similarly low gain in AUC ($0.03$) and a high discrepancy ($0.65$), coupled with a slightly reduced obscurity ($0.18$).

The influence of filtering methods is particularly striking. For data not subjected to filtering, represented by red triangles, the clustering of median points near the origin reflects limited changes in performance ($0.03$) and predictive multiplicity in terms of obscurity ($0.01$), and discrepancy ($0.2$). In contrast, correlation-based filtering amplifies these metrics for balancing methods such as ANSMOTE, where the average discrepancy rises to $0.6$, obscurity increases to $0.08$, and the gain in AUC performance reaches $0.11$. Significance-based filtering results in even higher predictive multiplicity for methods such as RSL-SMOTE, with average discrepancy values of $0.7$ and obscurity of $0.09$, although the performance gain ($0.08$) remains moderate.

The Kruskal-Wallis test tests the significance of increasing the model performance using balancing methods, followed by the Dunn pairwise posthoc comparison test to determine the insignificant changes. It showed that the balancing methods statistically improved the model performance ($\chi^2_{KW} (10) = 85.12, p < 0.0001$***) except Undersampling, Near Miss, RSLSMOTE, and SLSMOTE ($p_i > 0.05, i = 1, 2, 3, 4$). The KW test is also performed to test the significance of increasing the model performance using filtering methods, and the result showed no significant difference ($\chi^2_{KW} (2) = 0.39, p = 0.82$) between the using of filtering methods and the unfiltered (aka original) datasets.  

Considering the block effect, the Friedman test is used to check whether there is a significant difference between the performance gain of the balancing methods. The results for using balancing methods considering the filtering methods show that there is a significant performance gain from using the balancing methods ($\chi^2_{F} (10) = 20.556, p = 0.0044$**), the results for using filtering methods considering the balancing techniques show that there is no significant improvement on the model performances ($\chi^2_{F} (10) = 4.3226, p = 0.1152$). 

These results support the conclusion that balancing methods improve AUC performance to different degrees. SMOTE variants provide a middle ground between boosting performance and maintaining predictive multiplicity. Undersampling focuses more on increasing model disagreement and does not improve predictive performance. Meanwhile, filtering methods highlight the trade-offs involved, emphasizing the importance of carefully selecting methods to accomplish analytical goals.

\subsection*{\textbf{RQ6. Can complexity metrics be an indicator of predictive multiplicity?}}

The relationship between complexity metrics and disagreement metrics is examined to determine whether complexity can signal predictive multiplicity. To assess these relationships, Spearman's rank correlation coefficients \cite{Spearman_1987} and the corresponding significance tests \cite{Hollander_2013} were used. Table~\ref{tab:correlation} presents the correlation coefficients and the results of the significance tests. The findings indicate that certain complexity metrics show statistically significant relationships with obscurity and discrepancy.

\begin{table}[H]
    \centering     
    \scriptsize
    \caption{The correlation coefficients and the correlation test results between the complexity and the disagreement metrics}
    \label{tab:correlation}
    \begin{tabular}{crlrl}\hline
        \multirow{2}{*}{Complexity metric} &\multicolumn{2}{c}{obscurity}  &\multicolumn{2}{c}{discrepancy}\\\cline{2-5}
                            & $r$       & $p$-value             & $r$       & $p$-value \\\hline
        T2                  & -0.019    & 0.9332                & -0.070    & 0.7614\\ 
        T3                  & -0.109    & 0.6377                & -0.172    & 0.4546\\
        T4                  & -0.215    & 0.3485                & -0.292    & 0.1986\\\hline
        \cellcolor{yellow}L1         & \cellcolor{yellow}0.767     & \cellcolor{yellow}0.0001***            & \cellcolor{yellow}0.560     & \cellcolor{yellow}0.0082**\\
        \cellcolor{yellow}L2         & \cellcolor{yellow}0.754     & \cellcolor{yellow}0.0001***             & \cellcolor{yellow}0.577     & \cellcolor{yellow}0.0061**\\
        \cellcolor{yellow}L3         & \cellcolor{yellow}0.740     & \cellcolor{yellow}0.0001***             & \cellcolor{yellow}0.568     & \cellcolor{yellow}0.0071**\\\hline
        F1                  & -0.420    & 0.0587                & -0.334    & 0.1383\\
        F1v                 & -0.081    & 0.7241                & 0.195     & 0.3957\\
        F2                  & 0.094     & 0.6823                & 0.039     & 0.8644\\
        F3                  & 0.136     & 0.5542                & 0.013     & 0.9532\\
        F4                  & 0.302     & 0.1820                & 0.220     & 0.3360\\\hline
        N1                  & 0.619     & 0.0056**              & 0.447     & 0.0546\\
        N2                  & -0.019    & 0.9397                & 0.143     & 0.5591\\
        N3                  & 0.542     & 0.0180*               & 0.407     & 0.0835\\
        \cellcolor{yellow}N4         & \cellcolor{yellow}0.512     & \cellcolor{yellow}0.0266*               & \cellcolor{yellow}0.564     & \cellcolor{yellow}0.0118*\\
        T1                  & -0.147    & 0.5458                & -0.227    & 0.3494\\
        LSC                 & 0.438     & 0.0618                & 0.369     & 0.1195\\\hline
        \multicolumn{5}{r}{* $p$-value $<0.05$, ** $p$-value $<0.01$, *** $p$-value $<0.001$}
    \end{tabular}
\end{table}

For obscurity, the metrics L1 ($r = 0.767$), L2 ($r = 0.754$), and L3 ($r = 0.74$) exhibit strong positive correlations, while F1 ($r = -0.42$) demonstrates a moderate negative correlation. Additionally, N1 ($r = 0.619$), N3 ($r = 0.542$), and N4 ($r = 0.512$) exhibit significant positive correlations, with N4 also showing a moderate positive relationship with discrepancy ($r = 0.564$). The F2 metric, although not statistically significant in its correlation with obscurity ($r = 0.094$), shows no significant association with either metric.

For discrepancies, significant correlations were found with L1 ($r = 0.56$), L2 ($r = 0.577$), and L3 ($r = 0.568$), with moderate positive correlations. L1, L2, and L3 are particularly notable due to their strong positive relationships with both obscurity and discrepancy. N4, with a significant correlation in both metrics, also stands out, indicating its potential relevance for predicting multiplicity.

F1 shows a unique negative relationship with obscurity ($r = -0.42$), suggesting that as complexity increases in terms of F1, obscurity decreases. However, no significant correlation with discrepancy is observed for F1, F2, and F3, emphasizing their relative lack of association with predictive multiplicity in the dataset. Furthermore, the unique negative relationship between F1 and obscurity suggests that some complexity metrics may not always correlate positively with predictive multiplicity.

These results emphasize the relevance of linearity metrics (L1, L2, L3) and N4 in their correlation with the disagreement metrics, highlighting them as potential indicators of complexity.

\section{Conclusion}

In this paper, we have demonstrated the intricate dynamics between balancing and filtering methods, predictive multiplicity, and data complexity in imbalanced classification problems, emphasizing the principles of data-centric AI. The findings highlight the criticality of selecting tailored strategies that align with the complexity of the dataset to achieve robust and reliable ML models.

Balancing methods, especially ANSMOTE, inflate predictive multiplicity regarding discrepancy and obscurity metrics. For instance, while the original dataset has an average discrepancy of $0.4$ and obscurity of $0.1$, ANSMOTE increases these values to $0.6$ and $0.2$, respectively. Similarly, while impactful in some scenarios,  methods such as Near Miss and ADASYN demonstrate increased predictive multiplicity in high-complexity datasets, with discrepancy values rising to $0.7$ and obscurity to $0.2$. This underscores the need for judicious application of these methods, reflecting a data-centric AI principle of adapting strategies based on data characteristics.

Filtering methods, particularly the Significance Test, proved instrumental in reducing predictive multiplicity. For example, combining DBSMOTE with the Significance Test reduces the discrepancy from $0.5$ (without filtering) to $0.3$, and obscurity from $0.15$ to $0.1$. By focusing on statistically significant variables, this approach reduces variability in model predictions and enhances the reliability of a data-centric AI approach. Conversely, the absence of filtering leads to greater variability, with discrepancy values exceeding $0.6$ and obscurity rising above $0.15$.

Our findings further reveal that data complexity significantly moderates the effectiveness of these methods. In low-complexity datasets, RSL-SMOTE and BLSMOTE achieve minimal predictive multiplicity, with average discrepancy values around $0.3$ and obscurity near $0.05$. In contrast, Near Miss exacerbates multiplicity in high-complexity datasets, pushing discrepancy to $0.7$ and obscurity to $0.2$. These observations reinforce the importance of incorporating complexity metrics, such as L1, L2, L3, and N4, to inform the selection of balancing and filtering methods—a key principle of data-centric AI focusing on making decisions informed by the data itself.

Finally, the trade-off between predictive multiplicity and model performance must be navigated carefully. Balancing techniques like SMOTE and its derivatives offer a favorable compromise by enhancing performance while maintaining diversity within the Rashomon set. For example, SMOTE increases AUC by $0.1$ while keeping the discrepancy around $0.5$ and obscurity at $0.05$. In contrast, the Near Miss results in a lower AUC gain of $0.02$ but increases the discrepancy to $0.7$. Filtering methods, particularly those targeting variable significance, emphasize these benefits, enabling a refined balance between prediction accuracy and model consistency, aligning with the data-centric AI perspective of optimizing outcomes by improving data preprocessing and representation.

For practitioners addressing imbalanced classification problems, we recommend adopting a data-centric AI approach. This includes using balancing and filtering methods that consider the complexity of data. The results in this paper guide the choice of data-centric strategies. Adopting these strategies makes it possible to mitigate the challenges of predictive multiplicity while fostering reliable ML models. This reinforces the central role of data-centric methodologies in building robust AI systems tailored to the unique characteristics of the data.

\section*{Supplemental Materials}
The materials for reproducing the experiments and benchmark data sets can be found in the repository: \href{https://github.com/mcavs/data_centric_Rashomon_paper}{github.com/mcavs/data\_centric\_Rashomon\_paper}.
\section*{Acknowledgments}
This study is funded by the Polish National Science Centre under SONATA BIS grant 2019/34/E/ST6/00052 and Eskisehir Technical University Scientific Research Projects Commission under grant no. 24ADP116.

\section*{Declaration of generative AI and AI-assisted technologies in the writing process}

During the preparation of this paper, the author(s) used ChatGPT4.0 in order to grammar correction. After using this tool/service, the author(s) reviewed and edited the content as needed and take(s) full responsibility for the content of the publication.


\section*{Appendix}
\label{sec:appendix}

\subsection*{Data Complexity Measures}

This section summarizes various groups of metrics designed to assess the complexity of data sets. Data complexity measures assess sparsity and dimensionality, providing insight into how well the data can be represented and interpreted. Linearity measures focus on the ability of linear classifiers to separate classes effectively while overlapping measures assess the degree of class separability and the implications for classification performance. In addition, neighborhood measures analyze local relationships among data points, offering valuable insight into the distribution and connectivity of classes. Collectively, these metric groups enhance our understanding of data complexity, guiding the selection and application of appropriate modeling techniques.
\subsubsection*{Dimensionality Measures}

Dimensionality measures assess the size of the data set and the number of variables it contains. High-dimensional data can pose challenges because of the curse of dimensionality. Metrics such as the \textit{Average Number of Points per Dimension} ($T2$), \textit{Average Number of Points per PCA Component} ($T3$), and \textit{Ratio of Relevant PCA Components to Original Dimensions} ($T4$) given in this section will illustrate the effectiveness of dimensionality reduction techniques, providing insight into how to manage and interpret high-dimensional data effectively \cite{Lorena_et_al_2012}. The details of the dimensionality measures are given in Table~\ref{tab:dimensionality_measures}.

\begin{table}[ht]
\centering
\scriptsize
\caption{Dimensionality Measures}
\label{tab:dimensionality_measures}
\begin{tabular}{p{2cm}p{10cm}}
\toprule
\multicolumn{2}{l}{\textbf{\boldmath$T2$: Average Number of Points per Dimension}} \\ \midrule
\textbf{Formula} & $T2 = \frac{n}{p}$ \newline where $n$ is the number of observations, and $p$ is the number of dimensions. \\ 
\textbf{Description} & Reflects the sparsity of the data in the variable space. \\ 
\textbf{Interpretation} & Higher values suggest sufficient data points per dimension, indicating well-represented data and lower complexity. Lower values imply insufficient data points per dimension, leading to underrepresented dimensions and higher complexity. \\ \toprule

\multicolumn{2}{l}{\textbf{\boldmath$T3$: Average Number of Points per PCA Component}} \\ \midrule
\textbf{Formula} & $T3 = \frac{n}{k}$ \newline where $n$ is the number of observations, and $k$ is the number of PCA components capturing $95\%$ variance. \\ 
\textbf{Description} & Assesses data sparsity in the reduced PCA space. \\ 
\textbf{Interpretation} & Higher values suggest that fewer PCA components suffice to explain most of the variance, indicating lower complexity. Lower values imply that more PCA components are needed to represent the data, indicating higher complexity. \\ \toprule

\multicolumn{2}{l}{\textbf{\boldmath$T4$: Ratio of Relevant PCA Components to Original Dimensions}} \\ \midrule
\textbf{Formula} & $T4 = \frac{k}{p}$ \newline where $k$ is the number of relevant PCA components, and $p$ is the number of original dimensions. \\ 
\textbf{Description} & Represents the proportion of significant PCA components to original dimensions. \\ 
\textbf{Interpretation} & Higher values indicate a smaller subset of dimensions effectively represents the data, suggesting lower complexity. Lower values suggest a significant portion of dimensions is required for representation, leading to higher complexity. \\ \toprule
\end{tabular}
\end{table}

Table~\ref{tab:dataset_dimensionality} presents the metric values of the dimensionality complexity for the benchmark datasets in Table~\ref{tab:dataset}, columns T2, T3, and T4. Datasets \texttt{phoneme}, \texttt{yeast\_ml8}, and \texttt{mammography} exhibit higher T4 values, indicating greater complexity in that specific dimension, whereas others \texttt{MagicTelescope} and \texttt{spambase} show lower values between metrics. 

\begin{table}[ht]
    \centering     
    \scriptsize
    \caption{Dimensionality complexity metric values of the datasets}
    \begin{tabular}{lccc}
        \toprule
        Dataset                         & T2     & T3     & T4     \\\midrule
        \texttt{spambase}               & 0.0124 & 0.0004 & 0.0351 \\
        \texttt{MagicTelescope}         & 0.0005 & 0.0003 & 0.5000 \\
        \texttt{steel\_plates\_fault}   & 0.0170 & 0.0005 & 0.0303 \\
        \texttt{qsar-biodeg}            & 0.0388 & 0.0066 & 0.1707 \\
        \texttt{phoneme}                & 0.0009 & 0.0009 & 1.0000 \\
        \texttt{jm1}                    & 0.0019 & 0.0001 & 0.0486 \\
        \texttt{SpeedDating}            & 0.1145 & 0.0095 & 0.0833 \\
        \texttt{kc1}                    & 0.0100 & 0.0005 & 0.0476 \\
        \texttt{churn}                  & 0.0040 & 0.0002 & 0.0500 \\
        \texttt{pc4}                    & 0.0254 & 0.0007 & 0.0270 \\
        \texttt{pc3}                    & 0.0237 & 0.0006 & 0.0270 \\
        \texttt{abalone}                & 0.0019 & 0.0005 & 0.2500 \\
        \texttt{us\_crime}              & 0.0502 & 0.0160 & 0.3200 \\
        \texttt{yeast\_ml8}             & 0.0426 & 0.0306 & 0.7184 \\
        \texttt{pc1}                    & 0.0189 & 0.0009 & 0.0476 \\
        \texttt{ozone\_level\_8hr}      & 0.0284 & 0.0016 & 0.0556 \\
        \texttt{wilt}                   & 0.0010 & 0.0004 & 0.4000 \\
        \texttt{wine\_quality}          & 0.0022 & 0.0004 & 0.1818 \\
        \texttt{yeast\_me2}             & 0.0054 & 0.0047 & 0.8750 \\
        \texttt{mammography}            & 0.0005 & 0.0005 & 0.8333 \\
        \texttt{abalone\_19}            & 0.0019 & 0.0005 & 0.2500 \\ \bottomrule
    \end{tabular}
    \label{tab:dataset_dimensionality}
\end{table}

\subsubsection*{Linearity Measures}

Linearity measures evaluate how well the data can be separated using linear classification algorithms. These measurements are critical for understanding the linearity of relationships between classes and determining the suitability of linear models. Metrics like the \textit{Sum of Error Distances by Linear Programming} ($L1$), \textit{Error Rate of the Linear Classifier} ($L2$), and \textit{Non-Linearity of the Linear Classifier} ($L3$) will be discussed to assess the performance of linear classifiers in distinguishing between different classes \cite{Orriols_et_al_2010}. The details of the linearity measures are given in Table~\ref{tab:linearity_measures}.

\begin{table}[ht]
\centering
\scriptsize
\caption{Linearity Measures}
\label{tab:linearity_measures}
\begin{tabular}{p{2cm}p{10cm}}
\toprule
\multicolumn{2}{l}{\textbf{\boldmath$L1$: Sum of Error Distances by Linear Programming}} \\ \midrule
\textbf{Formula} & $L1 = \sum_{i=1}^{n} d(\textbf{x}_i, f(\textbf{x}_i))$ \newline where $n$ is the number of observations, and $d(\textbf{x}_i, f(\textbf{x}_i))$ is the distance between observation $\textbf{x}_i$ and the decision boundary. \\ 
\textbf{Description} & The sum of distances from misclassified points to the decision boundary. \\ 
\textbf{Interpretation} & Higher values indicate that misclassified examples are far from the boundary, suggesting high dataset complexity. Lower values imply that misclassified examples are close to the boundary, indicating lower complexity. \\ \toprule

\multicolumn{2}{l}{\textbf{\boldmath$L2$: Error Rate of the Linear Classifier}} \\ \midrule
\textbf{Formula} & $L2 = \frac{1}{n} \sum_{i=1}^{n} \mathbbm{1}[f(\textbf{x}_i) \neq y_i]$ \newline where $n$ is the number of observations, and $\mathbbm{1}[\cdot]$ is the indicator function (1 for misclassification, 0 otherwise). \\ 
\textbf{Description} & The proportion of misclassified examples by a linear SVM. \\ 
\textbf{Interpretation} & Higher values suggest the linear classifier struggles with accuracy, reflecting high dataset complexity. Lower values indicate effective classification by the linear model, showing lower complexity. \\ \toprule

\multicolumn{2}{l}{\textbf{\boldmath$L3$: Non-Linearity of the Linear Classifier}} \\ \midrule
\textbf{Formula} & $L3 = \frac{1}{n'} \sum_{i=1}^{n'} \mathbbm{1}[f(\textbf{x}_i^{\text{new}}) \neq y_i^{\text{new}}]$ \newline where $n'$ is the number of new examples, and $\textbf{x}_i^{\text{new}}$ are interpolated examples. \\ 
\textbf{Description} & The error rate of the linear classifier on interpolated examples. \\ 
\textbf{Interpretation} & Higher values indicate that interpolated examples are misclassified, suggesting significant non-linear relationships. Lower values imply good generalization to interpolated examples, indicating simpler, more linear relationships. \\ \toprule
\end{tabular}
\end{table}

Table~\ref{tab:dataset_linearity} presents three linearity metrics L1, L2, and L3 for the benchmark datasets. High values suggest greater linearity complexity: for example, the \texttt{MagicTelescope} dataset exhibits one of the highest linearity complexities across all metrics. On the other hand, the \texttt{steel\_plates\_fault} dataset displays extreme simplicity, with all metric values being zero, indicating no linearity complexity. Other datasets, such as \texttt{spambase} and \texttt{phoneme}, also show relatively high complexity, whereas \texttt{mammography} and \texttt{abalone\_19} present some of the lowest complexity measures, reflecting their closer alignment with linear separability.

\begin{table}[ht]
    \centering     
    \scriptsize
    \caption{Linearity complexity metric values of the datasets}
    \begin{tabular}{lccc}
        \toprule
        Dataset                         & L1     & L2     & L3     \\\midrule
        \texttt{spambase}               & 0.2591 & 0.1051 & 0.0875 \\
        \texttt{MagicTelescope}         & 0.3196 & 0.1533 & 0.1380 \\
        \texttt{steel\_plates\_fault}   & 0.0000 & 0.0000 & 0.0000 \\
        \texttt{qsar-biodeg}            & 0.2735 & 0.1127 & 0.0982 \\
        \texttt{phoneme}                & 0.3188 & 0.1560 & 0.1422 \\
        \texttt{jm1}                    & 0.2813 & 0.1401 & 0.1364 \\
        \texttt{SpeedDating}            & 0.2301 & 0.0912 & 0.0826 \\
        \texttt{kc1}                    & 0.2065 & 0.1021 & 0.0964 \\
        \texttt{churn}                  & 0.2161 & 0.0990 & 0.0918 \\
        \texttt{pc4}                    & 0.1730 & 0.0726 & 0.0672 \\
        \texttt{pc3}                    & 0.1591 & 0.0737 & 0.0691 \\
        \texttt{abalone}                & 0.1681 & 0.0772 & 0.0763 \\
        \texttt{us\_crime}              & 0.1227 & 0.0439 & 0.0383 \\
        \texttt{yeast\_ml8}             & 0.1345 & 0.0638 & 0.0624 \\
        \texttt{pc1}                    & 0.1177 & 0.0550 & 0.0527 \\
        \texttt{ozone\_level\_8hr}      & 0.1288 & 0.0487 & 0.0468 \\
        \texttt{wilt}                   & 0.1047 & 0.0432 & 0.0418 \\
        \texttt{wine\_quality}          & 0.0755 & 0.0333 & 0.0327 \\
        \texttt{yeast\_me2}             & 0.0778 & 0.0293 & 0.0278 \\
        \texttt{mammography}            & 0.0458 & 0.0167 & 0.0153 \\
        \texttt{abalone\_19}            & 0.0162 & 0.0075 & 0.0072 \\ \bottomrule
    \end{tabular}
    \label{tab:dataset_linearity}
\end{table}
\subsubsection*{Overlapping Measures} 

This section includes metrics that evaluate the overlap between different classes. Specifically, the extent to which classes overlap can significantly affect classification performance. Metrics such as \textit{Maximum Fisher’s Discriminant Ratio} ($F1$), \textit{Directional-Vector Maximum Fisher’s Discriminant Ratio} ($F1v$), \textit{Volume of the Overlapping Region} ($F2$), \textit{Maximum Individual variable Efficiency} ($F3$), and \textit{Collective variable Efficiency} ($F4$) provide the insights into the ability to distinguish of variables in the dataset, which can complicate classification tasks. Table~\ref{tab:overlapping_measures} gives the details of the overlapping measures.

\begin{table}[ht]
\centering
\scriptsize
\caption{Overlapping Measures}
\label{tab:overlapping_measures}
\begin{tabular}{p{2cm}p{10cm}}
\toprule
\multicolumn{2}{l}{\textbf{\boldmath$F1$: Maximum Fisher’s Discriminant Ratio}} \\ \midrule
\textbf{Formula} & $F1 = \max_j \left( \frac{(\mu_{j,1} - \mu_{j,2})^2}{\sigma_{j,1}^2 + \sigma_{j,2}^2} \right)$ \newline where $\mu_{j,c}$ is the mean of variable $j$ for class $c$, and $\sigma_{j,c}^2$ is the variance of variable $j$ for class $c$. \\ 
\textbf{Description} & Evaluates class separability for each variable. \\ 
\textbf{Interpretation} & Higher values indicate significant separation between classes, suggesting lower complexity. Lower values imply substantial overlap between classes, indicating higher complexity. \\ \toprule

\multicolumn{2}{l}{\textbf{\boldmath$F1v$: Directional-Vector Maximum Fisher’s Discriminant Ratio}} \\ \midrule
\textbf{Formula} & Not explicitly defined. \\ 
\textbf{Description} & Searches for a vector projection to maximize class separability. \\ 
\textbf{Interpretation} & Higher values suggest a direction exists where classes are effectively separated, implying lower complexity. Lower values indicate no suitable separation direction, reflecting higher complexity. \\ \toprule

\multicolumn{2}{l}{\textbf{\boldmath$F2$: Volume of the Overlapping Region}} \\ \midrule
\textbf{Formula} & $F2 = \sum_{j=1}^{p} \left( \min(\textbf{x}_{j,1}^{\max}, \textbf{x}_{j,2}^{\max}) - \max(\textbf{x}_{j,1}^{\min}, \textbf{x}_{j,2}^{\min}) \right)$ \newline where $\textbf{x}_{j,c}^{\max}$ and $\textbf{x}_{j,c}^{\min}$ are the maximum and minimum values of variable $j$ for class $c$. \\ 
\textbf{Description} & Quantifies the overlap in variable distributions between classes. \\ 
\textbf{Interpretation} & Higher values indicate significant overlap, suggesting high dataset complexity. Lower values reflect smaller overlaps, indicating lower complexity. \\ \toprule

\multicolumn{2}{l}{\textbf{\boldmath$F3$: Maximum Individual Variable Efficiency}} \\ \midrule
\textbf{Formula} & $F3 = \max_j \left( \frac{n_{\text{non-overlap}, j}}{n} \right)$ \newline where $n_{\text{non-overlap}, j}$ is the number of examples outside overlapping region for variable $j$, and $n$ is the total number of examples. \\ 
\textbf{Description} & Ratio of non-overlapping examples for each variable. \\ 
\textbf{Interpretation} & Higher values suggest a variable effectively distinguishes between classes, indicating lower complexity. Lower values imply limited discriminatory power of the variable, indicating higher complexity. \\ \toprule

\multicolumn{2}{l}{\textbf{\boldmath$F4$: Collective Variable Efficiency}} \\ \midrule
\textbf{Formula} & $F4 = \frac{n_{\text{discriminated}}}{n}$ \newline where $n_{\text{discriminated}}$ is the number of separated examples, and $n$ is the total number of examples. \\ 
\textbf{Description} & Measures how well variables collectively separate the data. \\ 
\textbf{Interpretation} & Higher values indicate effective collective discrimination, suggesting lower complexity. Lower values reflect poor collective discrimination, indicating higher complexity. \\ \toprule
\end{tabular}
\end{table}

Table~\ref{tab:dataset_overlapping} presents the overlapping complexity metric values for the benchmark datasets. The metric values help to assess the separability and overlap of the classes in the data sets, guiding the understanding of potential challenges in model training. For instance, datasets \texttt{yeast\_ml8} and \texttt{abalone\_19} demonstrate high F1 scores, while \texttt{steel\_plates\_fault} shows a significant F1v value indicating notable variance within a class. These metrics are crucial for assessing the data structure and informing pre-processing or model selection strategies.

\begin{table}[ht]
    \centering     
    \scriptsize
    \caption{Overlapping complexity metric values of the datasets}
    \begin{tabular}{@{}lccccc@{}}
        \toprule
        Dataset                         & F1     & F1v    & F2          & F3     & F4     \\\midrule
        \texttt{spambase}               & 0.9666 & 0.1581 & 2.5331e-33  & 0.9087 & 0.6153 \\
        \texttt{MagicTelescope}         & 0.9536 & 0.3189 & 0.0815      & 0.9941 & 0.9821 \\
        \texttt{steel\_plates\_fault}   & 0.9725 & 0.7603 & 0.0000      & 0.7929 & 0.0000 \\
        \texttt{qsar-biodeg}            & 0.9521 & 0.1854 & 0.0001      & 0.9203 & 0.6644 \\
        \texttt{phoneme}                & 0.9297 & 0.3864 & 0.2708      & 0.8777 & 0.8651 \\
        \texttt{jm1}                    & 0.9663 & 0.7567 & 2.4540e-09  & 0.9989 & 0.9928 \\
        \texttt{SpeedDating}            & 0.9825 & 0.1958 & 0.0003      & 0.9103 & 0.7395 \\
        \texttt{kc1}                    & 0.9010 & 0.3736 & 0.0010      & 0.9905 & 0.9592 \\
        \texttt{churn}                  & 0.9874 & 0.3538 & 0.1190      & 0.9942 & 0.9696 \\
        \texttt{pc4}                    & 0.9725 & 0.4140 & 1.7536e-14  & 0.8765 & 0.6495 \\
        \texttt{pc3}                    & 0.9831 & 0.6842 & 8.1264e-21  & 0.9354 & 0.8682 \\
        \texttt{abalone}                & 0.9339 & 0.4615 & 0.0021      & 0.8781 & 0.7939 \\
        \texttt{us\_crime}              & 0.9555 & 0.1064 & 0.0003      & 0.8325 & 0.1199 \\
        \texttt{yeast\_ml8}             & 0.9986 & 0.4935 & 1.3611e-16  & 0.9590 & 0.2718 \\
        \texttt{pc1}                    & 0.9583 & 0.4646 & 3.7336e-06  & 0.9468 & 0.7096 \\
        \texttt{ozone\_level\_8hr}      & 0.9745 & 0.2159 & 1.3477e-18  & 0.7447 & 0.2908 \\
        \texttt{wilt}                   & 0.9929 & 0.2207 & 6.1094e-05  & 0.8493 & 0.6524 \\
        \texttt{wine\_quality}          & 0.9941 & 0.3260 & 0.0027      & 0.9706 & 0.9296 \\
        \texttt{yeast\_me2}             & 0.9803 & 0.2024 & 0.0000      & 0.9084 & 0.6887 \\
        \texttt{mammography}            & 0.9441 & 0.0598 & 0.0051      & 0.9732 & 0.9491 \\
        \texttt{abalone\_19}            & 0.9970 & 0.3157 & 0.0008      & 0.6655 & 0.5190 \\\bottomrule
    \end{tabular}
    \label{tab:dataset_overlapping}
\end{table}
\subsubsection*{Neighborhood Measures} 

Neighborhood measures assess the local structures of the data points and the relationships between classes. Such metrics are useful for understanding the separation between classes and evaluating how well the model performs. Measurements such as \textit{Fraction of Borderline Points} ($N1$), \textit{Ratio of Intra-Class to Extra-Class Nearest Neighbor Distance} ($N2$), \textit{Error Rate of the 1-nearest Neighbor Classifier} ($N3$), \textit{Non-Linearity of the 1-nearest Neighbor Classifier} ($N4$) \cite{Ho_and_Basu_2002}, and \textit{Fraction of Hyperspheres Covering Data} ($T1$) \cite{Ho_et_al_2006}  and Local Set Average Cardinality (LSC) \cite{Leyva_et_al_2014} help in analyzing the effectiveness of classifiers based on neighborhood information. The details of the neighborhood measures are given in Table~\ref{tab:neighborhood_measures}.

\begin{table}[ht]
\centering
\scriptsize
\caption{Neighborhood Measures}
\label{tab:neighborhood_measures}
\begin{tabular}{p{2cm}p{10cm}}
\toprule
\multicolumn{2}{l}{\textbf{\boldmath$N1$: Fraction of Borderline Points}} \\\midrule
\textbf{Formula} & $N1 = \frac{n_{\text{borderline}}}{n}$ \newline where $n_{\text{borderline}}$ is the number of points connected to neighbors of a different class, and $n$ is the total number of examples. \\ 
\textbf{Description} & Fraction of points on the boundary between classes in the Minimum Spanning Tree (MST). \\ 
\textbf{Interpretation} & Higher values indicate a significant presence of borderline points, suggesting high complexity and overlapping classes. Lower values imply fewer borderline points, indicating well-separated classes and lower complexity. \\ \toprule

\multicolumn{2}{l}{\textbf{\boldmath$N2$: Ratio of Intra-Class to Extra-Class Nearest Neighbor Distance}} \\ \midrule
\textbf{Formula} & $N2 = \frac{\sum_{i=1}^{n} d(\textbf{x}_i, \textbf{x}_i^{\text{in}})}{\sum_{i=1}^{n} d(\textbf{x}_i, \textbf{x}_i^{\text{out}})}$ \newline where $d(\textbf{x}_i, \textbf{x}_i^{\text{in}})$ is the distance to the nearest neighbor of the same class, and $d(\textbf{x}_i, \textbf{x}_i^{\text{out}})$ is the distance to the nearest neighbor of a different class. \\ 
\textbf{Description} & Measures separation between classes. \\ 
\textbf{Interpretation} & Higher values suggest well-separated classes with lower complexity. Lower values indicate similar intra- and inter-class distances, suggesting higher complexity and overlapping classes. \\ \toprule

\multicolumn{2}{l}{\textbf{\boldmath$N3$: Error Rate of the 1-Nearest Neighbor Classifier}} \\ \midrule
\textbf{Formula} & $N3 = \frac{1}{n} \sum_{i=1}^{n} \mathbbm{1}[f(\textbf{x}_i^{\text{train}}) \neq y_i]$ \newline where $f(\textbf{x}_i^{\text{train}})$ is the predicted label, and $y_i$ is the true label. \\ 
\textbf{Description} & Error rate of 1NN classifier using leave-one-out cross-validation. \\ 
\textbf{Interpretation} & Higher values indicate poor classification performance, suggesting high complexity and overlapping patterns. Lower values suggest effective classification with lower complexity and well-separated classes. \\ \toprule

\multicolumn{2}{l}{\textbf{\boldmath$N4$: Non-Linearity of the 1-Nearest Neighbor Classifier}} \\ \midrule
\textbf{Formula} & $N4 = \frac{1}{n'} \sum_{i=1}^{n'} \mathbbm{1}[f(\textbf{x}_i^{\text{new}}) \neq y_i^{\text{new}}]$ \newline where $n'$ is the number of interpolated examples, $f(\textbf{x}_i^{\text{new}})$ is the predicted label for interpolated data, and $y_i^{\text{new}}$ is the true label for interpolated data. \\ 
\textbf{Description} & Performance of 1NN classifier on interpolated data points. \\ 
\textbf{Interpretation} & Higher values reflect significant nonlinearity and complex relationships in data. Lower values imply linearity and simplicity in data structure. \\ \toprule

\multicolumn{2}{l}{\textbf{\boldmath$T1$: Fraction of Hyperspheres Covering Data}} \\ \midrule
\textbf{Formula} & $T1 = \frac{n_{\text{remaining}}}{n}$ \newline where $n_{\text{remaining}}$ is the number of remaining hyperspheres, and $n$ is the total number of examples. \\ 
\textbf{Description} & Fraction of remaining hyperspheres after pruning smaller nested ones. \\ 
\textbf{Interpretation} & Higher values suggest that fewer hyperspheres are sufficient to cover the data, indicating lower complexity. Lower values indicate more hyperspheres are needed, reflecting high complexity and overlapping data. \\ \toprule

\multicolumn{2}{l}{\textbf{\boldmath$LSC$: Local Set Average Cardinality}} \\ \midrule
\textbf{Formula} & $\text{LSC} = \frac{1}{n} \sum_{i=1}^{n} |LS(\textbf{x}_i)|$ \newline where $LS(\textbf{x}_i)$ is the local set of example $\textbf{x}_i$. \\ 
\textbf{Description} & Measures the average size of local sets for points. \\ 
\textbf{Interpretation} & Higher values indicate well-defined neighborhoods, suggesting lower complexity. Lower values reflect poorly defined neighborhoods, suggesting higher complexity. \\ \toprule
\end{tabular}
\end{table}

Table~\ref{tab:dataset_neighborhood} presents the benchmark datasets along with their neighborhood complexity metrics. It includes six metrics: \( N1 \), \( N2 \), \( N3 \), \( N4 \), \( T1 \), and \( LSC \) for each dataset. The \texttt{spambase} dataset has a moderate \( N2 \) value of 0.2714 and an \( LSC \) value of 0.9939, indicating that the dataset exhibits relatively low complexity in its neighborhood structure. Conversely, the \texttt{kc1} dataset shows a lower \( LSC \) of 0.9816 and a higher \( N4 \) value of 0.2313, suggesting more challenging neighborhood characteristics. The \texttt{MagicTelescope} and \texttt{mammography} datasets have all metrics marked as \textbf{NA} in the table, indicating that these metrics were not computed for these datasets because of their large sizes.  

\begin{table}[ht]
    \centering     
    \scriptsize
    \caption{Neighborhood complexity metric values of the datasets}
    \begin{tabular}{@{}lcccccc@{}}
        \toprule
        Dataset                         & N1        & N2        & N3        & N4        & T1        & LSC      \\ \midrule
        \texttt{spambase}               & 0.1614    & 0.2714    & 0.0873    & 0.0397    & 0.0006    & 0.9939   \\
        \texttt{MagicTelescope}         & NA        & NA        & NA        & NA        & NA        & NA   \\
        \texttt{steel\_plates\_fault}   & 0.0453    & 0.2520    & 0.0092    & 0.0376    & 0.0017    & 0.9718   \\
        \texttt{qsar-biodeg}            & 0.2729    & 0.3514    & 0.1696    & 0.0663    & 0.0024    & 0.9896   \\
        \texttt{phoneme}                & 0.1952    & 0.2462    & 0.0906    & 0.1752    & 0.0006    & 0.9805   \\
        \texttt{jm1}                    & 0.3585    & 0.3515    & 0.2477    & 0.2189    & 0.0001    & 0.9992   \\
        \texttt{SpeedDating}            & 0.3416    & 0.4092    & 0.2290    & 0.0028    & 0.0015    & 0.9928   \\
        \texttt{kc1}                    & 0.2754    & 0.2359    & 0.2110    & 0.2313    & 0.0009    & 0.9816   \\
        \texttt{churn}                  & 0.2246    & 0.4393    & 0.1424    & 0.0376    & 0.0002    & 0.9964   \\
        \texttt{pc4}                    & 0.1989    & 0.2823    & 0.1262    & 0.0637    & 0.0020    & 0.9670   \\
        \texttt{pc3}                    & 0.1912    & 0.2845    & 0.1196    & 0.0722    & 0.0023    & 0.9774   \\
        \texttt{abalone}                & 0.1984    & 0.3443    & 0.1448    & 0.1065    & 0.0008    & 0.9842   \\
        \texttt{us\_crime}              & 0.1253    & 0.3880    & 0.0867    & 0.0130    & 0.0012    & 0.9278   \\
        \texttt{yeast\_ml8}             & 0.1870    & 0.4522    & 0.1311    & 0.0004    & 0.0004    & 0.9893   \\
        \texttt{pc1}                    & 0.1334    & 0.2562    & 0.0730    & 0.0541    & 0.0040    & 0.9586   \\
        \texttt{ozone\_level\_8hr}      & 0.1207    & 0.4006    & 0.0773    & 0.0213    & 0.0009    & 0.9657   \\
        \texttt{wilt}                   & 0.0824    & 0.2802    & 0.0473    & 0.0359    & 0.0011    & 0.9831   \\
        \texttt{wine\_quality}          & 0.0818    & 0.2127    & 0.0416    & 0.0242    & 0.0011    & 0.9838   \\
        \texttt{yeast\_me2}             & 0.0646    & 0.2763    & 0.0404    & 0.0175    & 0.0059    & 0.9002   \\
        \texttt{mammography}            & NA        & NA        & NA        & NA        & NA        & NA   \\
        \texttt{abalone\_19}            & 0.0220    & 0.2138    & 0.0158    & 0.0071    & 0.0051    & 0.9402   \\ \bottomrule
    \end{tabular}
    \label{tab:dataset_neighborhood}
\end{table}

\end{document}